\pdfoutput=1
\documentclass[11pt]{article}

\usepackage[]{acl}

\usepackage{times}
\usepackage{latexsym}
\usepackage{amsmath, array, multirow, graphicx}

\usepackage[T1]{fontenc}
\usepackage[utf8]{inputenc}

\usepackage{microtype}

\usepackage{inconsolata}

\title{From Quantity to Quality: Boosting LLM Performance with Self-Guided Data Selection for Instruction Tuning}

\author{ 
  \textbf{Ming Li}\textsuperscript{1,2}, 
  \textbf{Yong Zhang}\textsuperscript{1}, 
  \textbf{Zhitao Li}\textsuperscript{1}, 
  \textbf{Jiuhai Chen}\textsuperscript{2}, 
  \textbf{Lichang Chen}\textsuperscript{2},
  \textbf{Ning Cheng}\textsuperscript{1}, \\
  \textbf{Jianzong Wang}\textsuperscript{1*} , 
  \textbf{Tianyi Zhou}\textsuperscript{2*},
  \textbf{Jing Xiao}\textsuperscript{1}, 
  \\
  \textsuperscript{1}Ping An Technology (Shenzhen) Co., Ltd., China \textsuperscript{2}University of Maryland \\
 \{minglii, tianyi\}@umd.edu, jzwang@188.com
}

\begin{document}
\maketitle

\renewcommand{\thefootnote}{}
\footnotetext{This work was done when Ming Li interned at Ping An Technology (Shenzhen) Co., Ltd., China.}
\renewcommand{\thefootnote}{\arabic{footnote}}
\renewcommand{\thefootnote}{}
\footnotetext{*Correspond to: Jianzong Wang (jzwang@188.com) and Tianyi Zhou (tianyi@umd.edu).}
\renewcommand{\thefootnote}{\arabic{footnote}}

\begin{abstract}
In the realm of Large Language Models (LLMs), the balance between instruction data quality and quantity is a focal point. Recognizing this, we introduce a self-guided methodology for LLMs to autonomously discern and select cherry samples from open-source datasets, effectively minimizing manual curation and potential cost for instruction tuning an LLM. Our key innovation, the Instruction-Following Difficulty (IFD) metric, emerges as a pivotal metric to identify discrepancies between a model's expected responses and its intrinsic generation capability. Through the application of IFD, cherry samples can be pinpointed, leading to a marked uptick in model training efficiency. Empirical validations on datasets like Alpaca and WizardLM underpin our findings; with a mere $10\%$ of original data input, our strategy showcases improved results. This synthesis of self-guided cherry-picking and the IFD metric signifies a transformative leap in the instruction tuning of LLMs, promising both efficiency and resource-conscious advancements. Codes, data, and models are available.\footnote{\url{https://github.com/tianyi-lab/Cherry_LLM}}
\end{abstract}

\section{Introduction}

Large Language Models (LLMs) have revolutionized the landscape of artificial intelligence \cite{touvron2023llama, touvron2023llama2, penedo2023refinedweb, Scao2022BLOOMA1}. Notable models such as GPT-3 \cite{NEURIPS2020_1457c0d6} and GPT-4 \cite{openai2023gpt4} leverage extensive datasets and advanced training methodologies to exhibit high-level text understanding and generation capabilities \cite{liu2023mmc, liu2024large, NEURIPS2023_66178bea, sun2024itd, liu2024mitigating}. 
Instruction tuning \cite{wei2022finetuned, Longpre2023TheFC} is a method employed to refine the performance of LLMs by providing specific guidelines or instructions during the model's training phase. It operates by supplying LLMs with explicit training instructions to produce the corresponding outputs that are more congruent with the desired ones \cite{xu2024survey}. A well-formulated instruction or prompt provides essential contextual information, refining the model's capability to generate relevant and task-specific outputs \cite{alpaca, NEURIPS2022_b1efde53}. 

% Based on the findings from \citet{wang2022super} and Self-Instruct \cite{wang-etal-2023-self-instruct} early experiments, reducing the number of instances per task does not degrade the model’s generalization performance to unseen tasks. 
Previously, instruction tuning was regarded as relying on amassing vast datasets \cite{khashabi-etal-2020-unifiedqa, ye-etal-2021-crossfit, wei2022finetuned, wang-etal-2022-super}. However, a seminal revelation from the LIMA \cite{zhou2023lima} highlights the art of instruction tuning: rather than a sheer volume of data, it's the quality of the data that dictates the model's performance. LIMA's findings emphasize that even a limited amount of manually curated, high-quality data can elevate the model's instruction-following capability. While it underscores the significance of data quality, the question of how to automatically identify high-quality data from a vast ocean of available datasets remains under investigation.

In our study, we introduce a novel approach for autonomously identifying the most impactful training samples, which we refer to as ``cherry data'', from extensive open-source datasets. These data samples are particularly effective in enhancing LLM instruction tuning. Central to our hypothesis is the idea that LLMs, through initial training with a small amount of instruction data, can inherently learn to discern and follow instructions, allowing them to estimate the difficulty of instruction data.

Our method involves a self-guided process that begins with familiarizing the model with a small subset of the dataset during the ``Learning from Brief Experience'' phase. This phase lays the groundwork for the subsequent ``Evaluating Based on Experience'' phase, where we introduce the Instruction-Following Difficulty (IFD) score. This metric evaluates how much help the instruction provides to the generation of the corresponding response, by comparing the loss in model responses with and without instructional context. The higher IFD score, indicating less instructional help, suggests a greater difficulty with instructions. On
the contrary, the lower IFD score represents that the
given instruction can directly benefit the language
model largely even without further training, representing the easiness and necessity of the instruction. Thus in the final ``Retraining from Self-Guided Experience'' phase, we use data with relatively large IFD scores as the cherry data to train our model, resulting in what we term ``cherry models''. This methodology, which emphasizes data quality over quantity, differs markedly from existing techniques that rely on external models for data curation.

\begin{figure}[t]
\centering 
\includegraphics[width=0.48\textwidth]{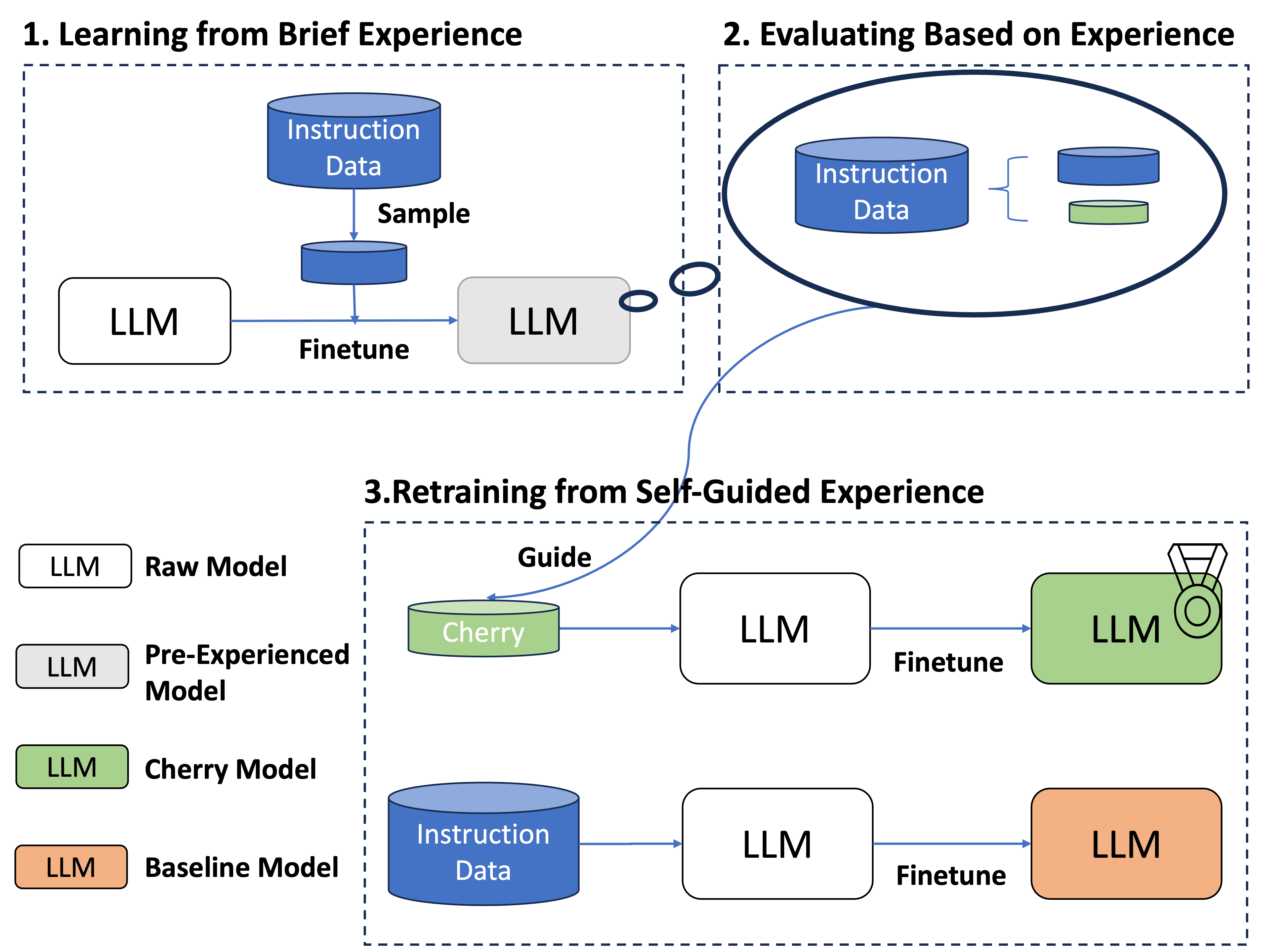} 
\caption{
Overview of our proposed method. 
} 
\label{model_structure} 
\vspace{-5mm}
\end{figure}

Extensive experimental results validate the efficacy of our method. By applying our methodology to the Alpaca and WizardLM instruction tuning datasets, our model outperforms the official Alpaca model with only approximately $5\%$ data selected and outperforms the reimplemented WizardLM model with approximately $10\%$ data selected. 
The key contributions of this paper:
\vspace{-2mm}
\begin{itemize}
    \item  We propose a self-guided approach enabling models to autonomously select the ``cherry data'' from vast open-source datasets. This innovation minimizes manual curation and optimizes the use of existing data resources, reducing costs and streamlining training. \vspace{-2mm}
    \item We introduce the Instruction-Following Difficulty (IFD) score as a metric to measure how much help the instruction can provide to the generation of the corresponding response, revealing the model-specific difficulty of the given data sample. Using the IFD metric, we can pinpoint the data that is most helpful for a specific model. \vspace{-2mm}
    \item Backed by validation on training datasets like Alpaca and WizardLM, our strategy demonstrates enhanced outcomes with only 10\% of the original data input, emphasizing our approach's efficiency and transformative impact.\vspace{-2mm}
    \item We provide a different model-specific view in measuring the difficulty of new instructions, which may benefit future instruction data generation work. \vspace{-2mm}
\end{itemize}

\vspace{-2mm}

\section{Methodology}

As illustrated in Figure \ref{model_structure}, our methodology is divided into three core phases: Learning from Brief Experience, Evaluating Based on Experience, and Retraining from Self-Guided Experience. The initial phase emphasizes equipping the model with a basic instruction-following capability. The subsequent phase introduces a novel metric to evaluate the instruction-following difficulty score of each sample based on the previously trained pre-experienced model. Finally, after obtaining difficulty scores in the target dataset, the cherry samples are selected to train our final model, which we call the cherry models. 

\subsection{Learning from Brief Experience}

This phase aims to equip the initial model with a basic instruction-following capability by forcing the model to first experience a subset of the target dataset. Specifically, for the initial full target dataset, $D_0$ contains $n$ triplets $x = (Instruction, [Input], Answer)$, we define the string $Question = map(Instruction, [Input]) $ as the complete instruction. The $map$ function is aligned with the original target dataset. Each word in $Question (Q)$ and $Answer (A)$ is denoted as $x^Q_i$ and $x^A_i$ respectively. Let $LLM_\theta$ denote the LLM we use and $\theta$ represent the weight of LLMs, specifically, $\theta_0$ represents the pre-trained base LLM model. Then the instruction embeddings for each sample $x_j$ are obtained by:
\begin{equation}
[h^Q_{j,1},..h^Q_{j,m}] = LLM_{\theta_0} (w^Q_{j,1},..w^Q_{j,m}) 
\end{equation}
\begin{equation}
h^Q_j = \frac{\sum^m_{i=1}h^Q_{j,i}}{m}
\end{equation}
where $w^Q_{j,i}$ represents the $i_{th}$ word of $Question$ strings of sample $j$ and $h^Q_{j,i}$ represents its corresponding last hidden states. 

To ensure the diversity of instructions exposed to the initial model, the basic clustering technique K-Means on these instruction embeddings is utilized. Motivated by LIMA's finding, we try to make this experience process as brief as possible by sampling only a few instances in each cluster which we call pre-experienced samples. Specifically, we generate $100$ clusters on instruction embeddings and sample $10$ instances in each cluster. Then the initial model is trained for only $1$ epoch with these samples to obtain our brief pre-experienced model.

\vspace{-1.2mm}%Put here to reduce too much white space after your table 

\subsection{Evaluating Based on Experience }

In this stage, we introduce the Instruction-Following Difficulty (IFD) score, a metric devised to evaluate the difficulty each instructional sample presents. Our primary motivation, adhering to the goal of minimizing cross-entropy loss in model training, guides the use of this metric. It specifically targets gauging the impact of training data by isolating the instructional component's influence from that of the answer. To achieve this, we employ a method that compares the loss when the model generates responses both with and without the context provided by instruction. This comparison is crucial as it forms the basis of the IFD score, effectively quantifying the extent to which instruction aids in response generation. 

In the instruction-tuning process, the loss of a sample pair $(Q, A)$ is calculated by continuously predicting the next tokens given the instruction $Q$ and their proceeding words:

\vspace{-3.2mm}%Put here to reduce too much white space after your table 

{
\small
\begin{equation}
L_\theta(A|Q) = -\frac{1}{N}\sum_{i=1}^{N} \log P(w^A_i |Q,  w^A_{1}, w^A_{2}, \ldots, w^A_{i-1}; \theta)
\end{equation}
}
where $N$ is the number of words of the ground-truth answer $A$. We denote this averaged cross-entropy loss as the \textbf{Conditioned Answer Score} $s_\theta(A|Q) = L_\theta(A|Q)$. 
This metric evaluates the model's capability to generate appropriate responses based on provided instructions. It measures the extent to which the model's output aligns with both the instruction and the corresponding correct answer. 

However, a higher $s_\theta(A|Q)$ does not mean a harder instruction to follow, it may be caused by the inherent characteristic of string $A$ itself. In the pre-LLM era, when models are required to learn both the knowledge and instruction-following ability during finetuning, it is reasonable to use $s_\theta(A|Q)$ as an indicator for the difficulty of a sample. However, things change a little for current LLMs, which have learned most of the knowledge in the pre-training phase and only need to learn to align and follow the instructions. Thus to estimate the difficulty of following instructions of a given sample, we introduce the \textbf{Direct Answer Score} $s_\theta(A)$:

\vspace{-4mm}%Put here to reduce too much white space after your table 

\begin{equation}
s_\theta(A) = -\frac{1}{N}\sum_{i=1}^{N} \log P(w^A_i |w^A_{1}, \ldots, w^A_{i-1}; \theta).
\end{equation}
which measures LLM's ability to generate this answer alone. It gauges the inherent difficulty or challenge posed by the answer in isolation, without the contextual guidance from its corresponding instruction. A higher direct answer score may suggest that the answer is inherently more challenging or intricate for the model to generate. 

Further, analyzing the balance between a sample's inherent challenge and the model's capabilities in following it sheds light on the intricacies of estimating the difficulty of the instruction of a given sample. Specifically, we try to estimate the \textbf{Instruction-Following Difficulty (IFD)} scores $\text{IFD}_\theta{(Q,A)}$ on following instruction of a given $(Q,A)$ pairs by calculating the ratio between $s_\theta(A)$ and $s_\theta(A|Q)$: 

\vspace{-4mm}%Put here to reduce too much white space after your table 

\begin{align}
\text{IFD}_\theta{(Q,A)} &= \frac{s_\theta(A|Q)}{s_\theta(A)} 
\end{align}

By utilizing this metric, the influence of LLM's intrinsic ability to fit the answer string is partially alleviated. The score measures the degree to which given instruction benefits the alignment of the corresponding response. High IFD scores infer the inability of the model to align responses to the given corresponding instructions, which in turn indicates the difficulty of an instruction. It is worth noting that this $\text{IFD}_\theta{(Q, A)}$ is a model-specific value, and we use our pre-experienced model to obtain all these values in the target dataset. 

To further filter out the sample whose instruction is misaligned with its response, a threshold of $1$ is set. Typically, the Conditioned Answer Score is always smaller than the Direct Answer Score due to the intrinsic nature of the next token prediction: With the context given, the prediction for the latter tokens should be easier. Thus if the IFD score is greater than $1$, the Conditioned Answer Score is even larger than the Direct Answer Score, which means the given instruction provides no useful context for the prediction of the response. In this situation, we think there exists a misalignment between the instruction and the corresponding response. 

Although our experiments reveal that learning from brief experiences is important, it makes the whole pipeline complicated and efficient. However, Superfiltering \cite{Li2024SuperfilteringWD} expands the use of IFD scores and shows that (1) Good prompting can relieve the burden of training a pre-experienced model; (2) The IFD scores calculated by weak language models are consistent with strong models, making it possible to utilize small models for filtering, further pushing forward the efficiency of data filtering for instruction tuning. 

\vspace{-1.2mm}%Put here to reduce too much white space after your table 

\section{Experimental Setup}

\subsection{Datasets}

\noindent
\textbf{Training Datasets}
The Alpaca dataset \cite{alpaca} encompasses $52002$ instruction-following samples. Developed using the self-instruct \cite{wang-etal-2023-self-instruct} approach with text-davinci-003. Though initially competitive, its dependence on text-davinci-003 posed data quality concerns. WizardLM dataset \cite{xu2023wizardlm} leverages the Evol-Instruct algorithm to improve the quality of instruction data. The incorporation of ChatGPT during the reformulation guarantees high fidelity of data. We utilize the WizardLM70K for our experiment. 

\noindent
\textbf{Test Datasets}
To ensure comprehensive and unbiased assessment, we employed $5$ diverse test sets: Vicuna \cite{vicuna2023}, Koala \cite{vu2023koala}, WizardLM \cite{xu2023wizardlm}, Self-instruct \cite{wang-etal-2023-self-instruct}, and LIMA \cite{zhou2023lima}. These test sets contain approximately $1000$ human curated instructions, open-domain or closed-domain for different tasks from different sources. Among them, Vicuna and WizardLM further provide the specific sub-category for each instruction, making it possible for in-depth analysis. 

\subsection{Implementation Details}

For experiments on the LLaMA-7B pre-trained model, our training configuration aligns with the original Alpaca and WizardLM, by utilizing the the Alpaca codebase\footnote{\url{https://github.com/tatsu-lab/stanford_alpaca}}. 
For experiments on LLaMA2-7B and LLaMA2-13B models, we utilize the Vicuna codebase\footnote{\url{https://github.com/lm-sys/FastChat}}. The detailed training configuration can be found in Appendix \ref{imlement}. 

\vspace{-1.2mm}%Put here to reduce too much white space after your table 

\subsection{Evaluation Metrics}

\subsubsection{Pair-wise Comparison}

Evaluating the instruction-following capabilities of LLMs is challenging. Extensive research is still dedicated to creating automated evaluation metrics for LLMs \cite{chang2023survey} since human evaluation is both labor-intensive and potentially influenced by subjective biases. Leveraging the recent advancements in independent LLM evaluations \cite{zheng2023judging, vicuna2023, alpaca_eval}, we utilize GPT4 and ChatGPT for comparative evaluations. Specifically, for each instruction in the test dataset, models that need to be compared are prompted to generate responses respectively. Then an API model, either GPT4 or ChatGPT, assigns scores for their responses. The model is regarded to be better in this dataset only if its answer is preferred by the judging model. 

In the evaluation, each model's response is rated by the judge on a scale from \(1\) to \(10\), reflecting attributes like relevance and accuracy. To further address the positional bias \cite{ko-etal-2020-look, wang2023large}, we send the responses of two models to the judge twice with different orders and compare their scores. Thus we define one model to be seen as winning only if it does not lose in both the ordering\footnote{Code, prompt, and testing dataset are provided: \url{https://github.com/tianyi-lab/Cherry_LLM} }, specifically: 
\vspace{-2mm}
\begin{itemize}
    \item \textbf{Wins:} outperforms in both or wins in one and ties in the other. \vspace{-3mm}
    \item \textbf{Tie:} ties in both or wins in one and loses in the other. \vspace{-3mm}
    \item \textbf{Loses:} lags in both or ties in one and loses in the other. \vspace{-3mm}
\end{itemize}

\subsubsection{Benchmarks}

The performances on two recently popular benchmarks for LLMs are also provided: Huggingface Open LLM Leaderboard and AlpacaEval Leaderboard. Huggingface Open LLM Leaderboard evaluates LLMs using \cite{eval-harness}, a unified framework to test generative language models on a large number of different evaluation tasks, on $4$ key benchmarks including ARC \cite{clark2018think}, HellaSwag \cite{zellers-etal-2019-hellaswag}, MMLU \cite{hendrycks2021measuring} and TruthfulQA \cite{lin-etal-2022-truthfulqa}. AlpacaEval Leaderboard provides an LLM-based automatic evaluation based on AlpacaFarm \cite{dubois2023alpacafarm} evaluation set, in which the model responses are compared with responses of Davinci003 by GPT4. 

\subsubsection{Human Evaluation}

To better illustrate the efficacy of our method, further human evaluation is conducted. Specifically, we randomly sampled $20$ instructions from each test set to generate a new random set containing $100$ instructions in total. Then $3$ human participants are asked to compare the responses generated by the models to be compared. For each comparison, 3 options are given (Win, Tie, and Loss) and the final results are determined by the majority voting of the participants. 

\vspace{-1.2mm}%Put here to reduce too much white space after your table 
\section{Experimental Results}

\subsection{Main Results}

\begin{figure}[t]
\centering 
\includegraphics[width=0.5\textwidth]{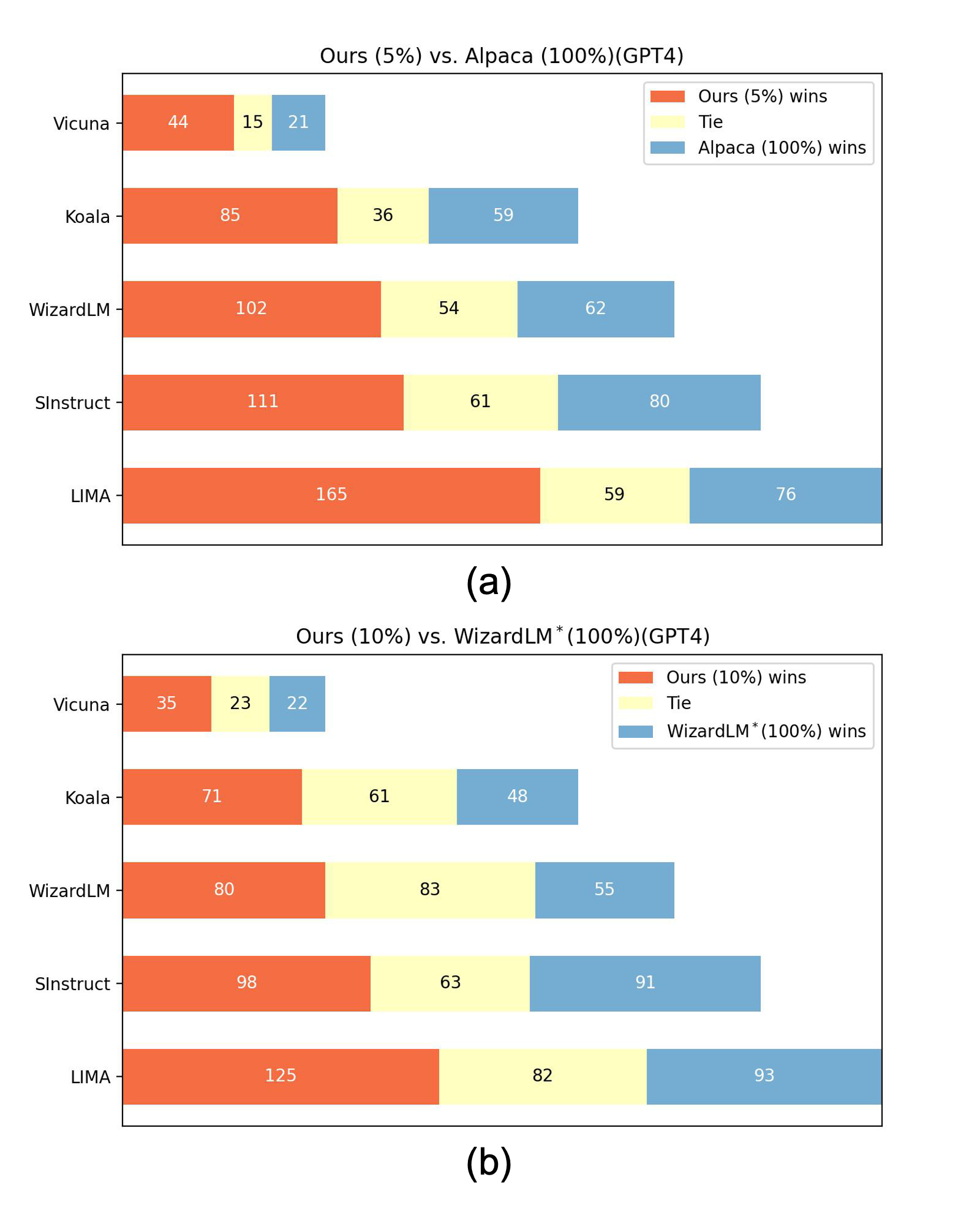} 
\vspace{-7mm}%Put here to reduce too much white space after your table 

\caption{
Comparing our models trained on selected data with full data. (a) Comparison between our model with $5\%$ Alpaca data and the official Alpaca model. (b) Comparison between our model with $10\%$ WizardLM data and the reimplemented WizardLM model. Both (a) and (b) use GPT4 as the judge. Each horizontal bar represents a comparison in a specific test set. 
} 
\label{main_result_chat} 
\vspace{-4mm}%Put here to reduce too much white space after your table 
\end{figure}

In this section, we first present our primary \textbf{pair-wise evaluation results} in Figure \ref{main_result_chat}. (a) our model trained with only approximately $5\%$ of the original Alpaca data beats the Alpaca model trained with full data. (b) our model trained with only approximately $10\%$ of the original WizardLM data beats the reimplemented WizardLM model under the same training configuration which is described in the Implementation Details.

\begin{figure}[t]
\centering 
\includegraphics[width=0.5\textwidth]{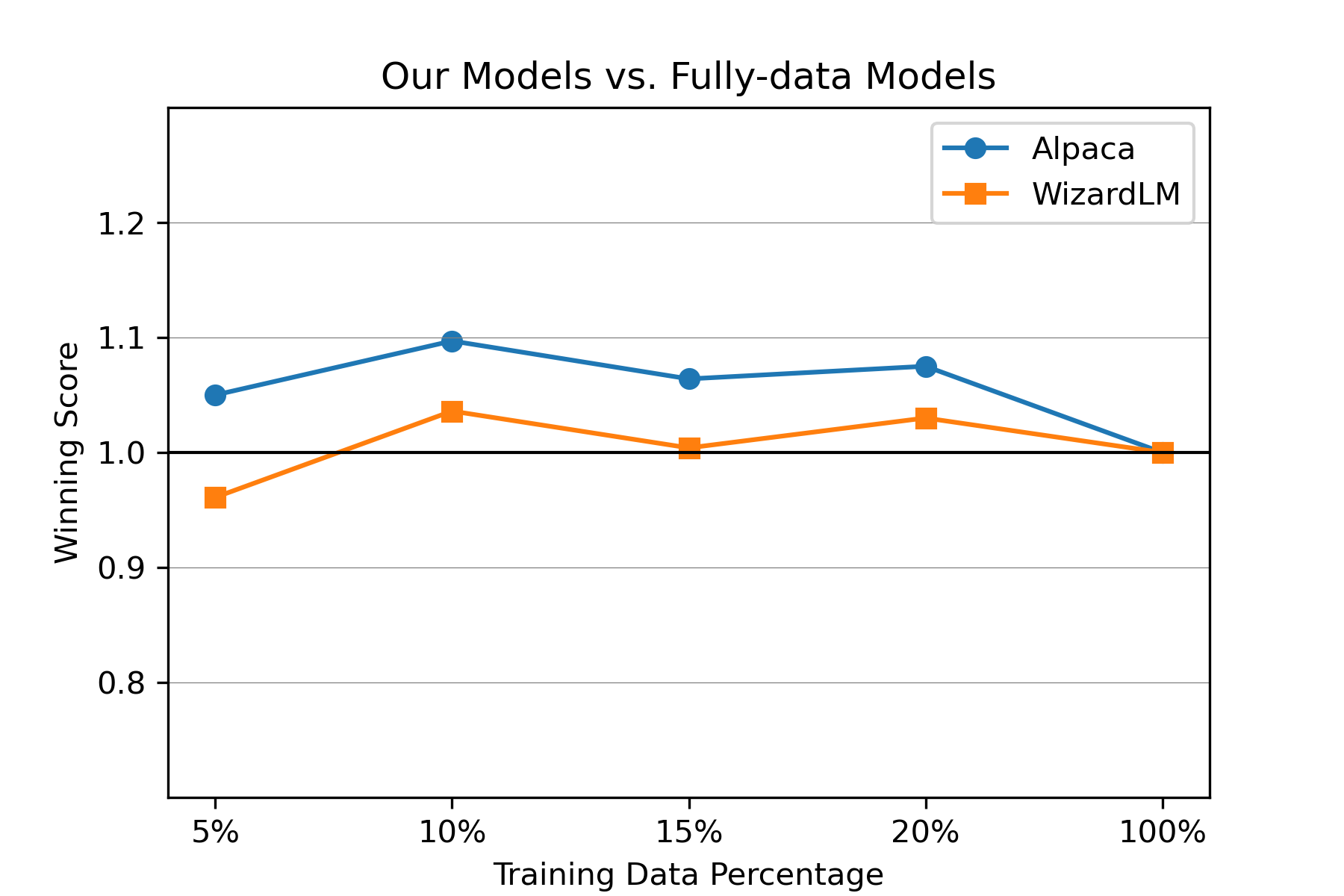} 
\vspace{-5.2mm}%Put here to reduce too much white space after your table 
\caption{
The winning score changes over data growth by comparing our models with full-data models. The winning score is calculated as (Num(Win)$-$Num(Lose))$/$Num(All) $+ 1$. The Number of Wins, Losses, and All are calculated across all five test sets we used. When the value is higher than $1.0$, it means this model performs better than the comparison. 
} 
\label{exp_2} 
\vspace{-4.2mm}%Put here to reduce too much white space after your table 
\end{figure}

Moreover, we craft subsets containing the top \(5\%\), \(10\%\), \(15\%\), and \(20\%\) of the training datasets to train models, enabling us to investigate the performance changes. As shown in Figure \ref{exp_2}, we draw the overall winning rate changes across the data growth, which is calculated as (Num(Win)$-$Num(Lose))$/$Num(All) $+ 1$, providing a direct indicator on the comparison with the full-data trained models. A consistent observation across both datasets is that with merely $10\%$ of selectively chosen data, our models manage to exceed the results of models trained on the full dataset. These findings not only highlight the efficiency of our data selection strategy but also underscore the potential of training powerful models with significantly reduced data requirements. By validating our approach on the renowned Alpaca dataset and the more intricate WizardLM dataset, the wide applicability and robustness of our proposed method are highlighted.

The comparison between our cherry models with baseline models on \textbf{Huggingface Open LLM Leaderboard and AlpacaEval Leaderboard} are presented in Table \ref{tbl:bench} where we can see our cherry model using $5\%$ Alpaca data outperforms the official Alpaca on both benchmarks, our cherry model using $10\%$ WizardLM data has a close performance compared with our re-implemented WizardLM. These results further showcase the effectiveness of our automatically selected data.
% Another comparison that is worth noting is the comparison between our reimplemented WizardLM and our cherry model with $40\%$ WizardLM data, under the same training configuration, our cherry model outperforms it with much less data. 

Moreover, the \textbf{human evaluation} results also showcase the usefulness of our method. When comparing the Cherry Alpaca (5\%) and the Alpaca (100\%), there are $49/100$ wins for our cherry alpaca, $25/100$ ties, and $26/100$ losses. When comparing the Cherry WizardLM (10\%) and the reimplemented WizardLM (100\%), there are $37/100$ wins for our Cherry WizardLM, $32/100$ ties, and $31/100$ losses. 

\begin{table*}[!tbh]
\centering
\scalebox{0.80}{
\begin{tabular}{l|ccccc|c}
\hline
& \multicolumn{5}{c|}{Huggingface Open LLM Leaderboard} & AlpacaEval \\
& Average & ARC & HellaSwag & MMLU & TruthfulQA & AlpacaEval \\
\hline
Official Alpaca & 50.21 & 42.65 & 76.91 & 41.73 & 39.55 & 26.46 \\
Ours (5\% Alpaca) & 52.06 & 53.92 & 79.49 & 36.51 & 38.33 & 34.74 \\
\hline
Reimplemented WizardLM$^*$ & 52.79 & 53.07 & 77.44 & 37.75 & 42.90 & 61.99 \\
% Ours (40\% WizardLM) & 52.83 & 53.07 & 77.79 & 35.29 & 45.17 & 65.09 \\
Ours (10\% WizardLM) & 51.59 & 52.90 & 78.95 & 33.08 & 41.41 & 61.44 \\
\hline
\end{tabular}
}
\caption{
The comparison of performance on Huggingface Open LLM Leaderboard and AlpacaEval Leaderboard. 
}
\label{tbl:bench}
\vspace{-6.2mm}
\end{table*}

% \vspace{-8.2mm}

\begin{figure}[t]
\centering 
\includegraphics[width=0.5\textwidth]{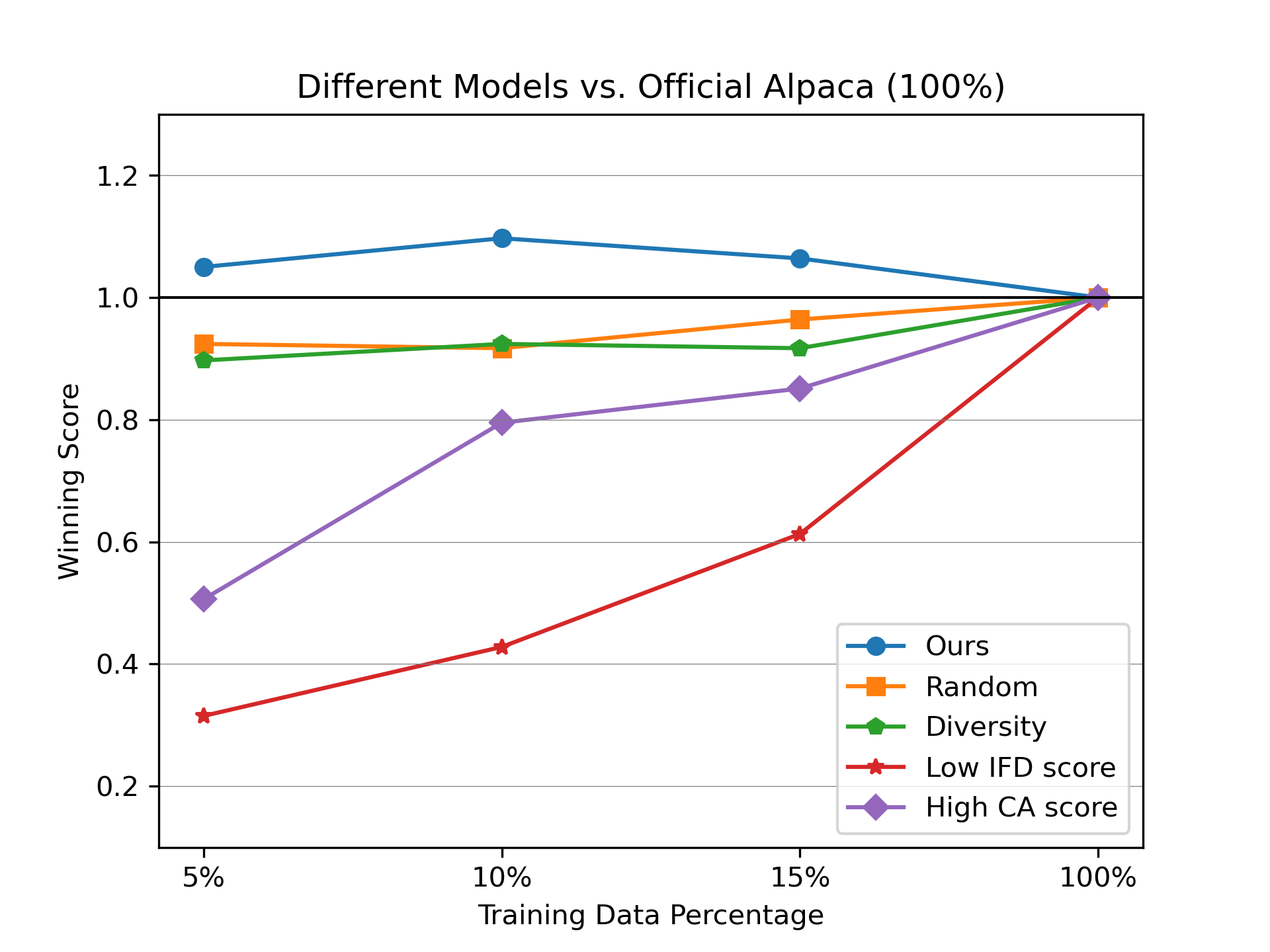} 
\caption{
The overall winning score changes by comparing models using different data selection strategies with the official Alpaca model. 
} 
\label{exp_4} 
\vspace{-4.2mm}%Put here to reduce too much white space after your table 
\end{figure}

\subsection{Ablation on Data Selection Mechanism}

In this section, we perform ablation studies comparing our method with other data selection mechanisms. The results are presented in Figure \ref{exp_4}, and the evaluation is based on ChatGPT as the judge on all $5$ test datasets. Moreover, the ablation results on the Open LLM Leaderboard and human evaluation are in Appendix \ref{ablation_data}.

\subsubsection{Data Randomly Selected}

We train various LLaMA-7B models using randomly chosen data and compare their performance with the model trained with full data. As shown in Figure \ref{exp_4} (labeled as \textit{Random}), models trained on $5\%$, $10\%$, or $15\%$ random data consistently underperformed against the official Alpaca model. Notably, with an equivalent amount of data, our model surpasses the performance of models using randomly selected data, underlining our method's superiority.

\subsubsection{Data with Diversity}

In this experiment, we train a series of models only considering the diversity of the data samples. Specifically, we utilize the K-means algorithm for the clustering, and then sample data from each cluster. It is a direct baseline for the situation where only the diversity of data is considered. As illustrated in Figure \ref{exp_4} (labeled as \textit{Diversity}), these models render subpar performance and are similar to the random trained models. This result shows that filtering data by only diversity is not enough for instruction tuning. 
% highlights the necessity of using difficult samples over pure diverse samples. 

\subsubsection{Data with Low IFD Score}

In this experiment, we aim to further underscore the efficacy of our proposed IFD score. We train models using data chosen based on low IFD scores on the pre-experienced model, a direct antithesis to our primary experimental setting. As illustrated in Figure \ref{exp_4} (labeled as \textit{Low IFD score}), models trained using low IFD scores obtain the least performance compared with all the methods. This observation highlights the prowess of our metric in sifting through high-quality data: a higher score consistently yields superior results, while a lower score deteriorates the model's intrinsic performance. This experiment directly showcases the consistent relationship between the performance and the IFD score values. 

\vspace{-6.2mm}

\subsubsection{Data with High CA Scores}

For this comparison, we train models on data selected by higher Conditioned Answer scores which is \textbf{equivalent to the loss or perplexity}, and is a commonly accepted baseline. As Figure \ref{exp_4} (labeled as \textit{High CA score}) elucidates, models in this group trail the official Alpaca model significantly. The salient difference between these models and ours rests on the elimination of Direct Answer scores. In models relying solely on CA scores, the underlying comprehension of the pre-trained LLM towards original answer texts isn't factored in, rendering high CA scores ineffective in gauging the intricate nuances of the instruction following.

\subsection{Ablation on Pre-Experienced Data}

\subsubsection{Number of Pre-Experience Data}

Following the findings from LIMA that $1000$ high-quality samples are enough to train a reasonably good model, we set the amount of data used for our pre-experienced model as $1000$. However, it is still under-investigated how many data samples are required to equip the model with basic instruction-following ability. Thus this section analyzes the necessity of employing pre-experience models and how the number of pre-experienced data affects the final performance of our cherry models. For these comparisons, we conduct the experiments where $0$, $100$, $300$, and $500$ pre-experienced samples are utilized to train the pre-experienced models. Using $0$ pre-experienced samples represents direct using the initial raw model as the pre-experienced model. We calculate the IFD scores from these different pre-experienced models and select the top $5\%$, $10\%$, and $15\%$ samples for training while keeping other experimental conditions constant. 

\begin{figure}[!htbp]
\centering 
\includegraphics[width=0.5\textwidth]{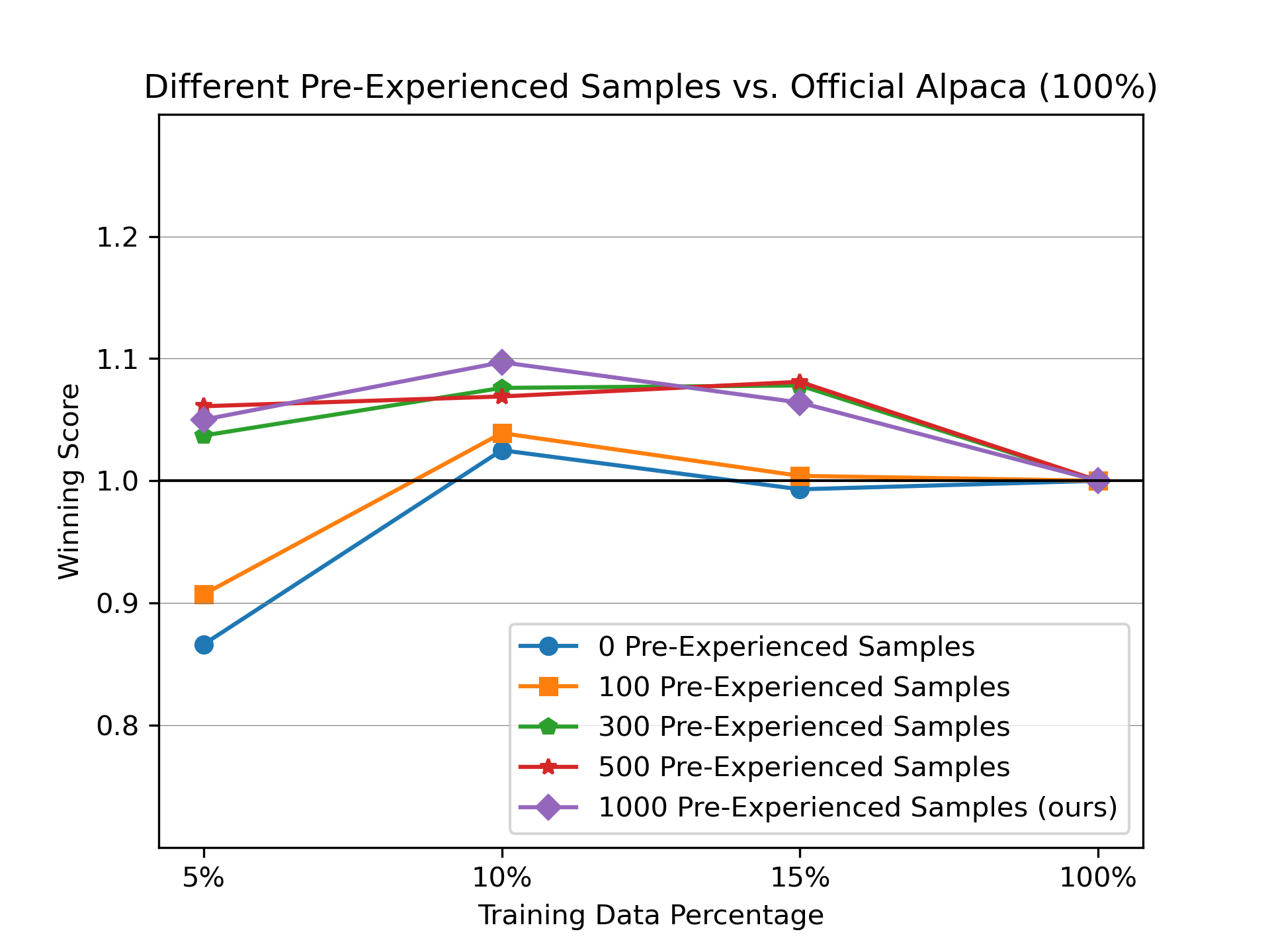} 
\caption{
The overall winning score changes by comparing models with different numbers of pre-experienced samples with the official Alpaca model. 
} 
\label{exp_pre} 
\vspace{-6.2mm}%Put here to reduce too much white space after your table 
\end{figure}

As shown in Figure \ref{exp_pre}, when no pre-experienced model is utilized, the corresponding cherry models have the least performance. However, even in the absence of a pre-experienced model, our IFD score remains effective in identifying the good training data subset as it outperforms the Alpaca model when using $10\%$ of the data. When $100$ samples are utilized, the corresponding cherry models are slightly better than no samples used but with a similar trend, which indicates that $100$ samples are not enough for the model to acquire the basic instruction-following ability. When adding the number of pre-experienced samples to $300$, a distinct performance gain is discovered, and further addition of samples does not make the performance of corresponding cherry models better. We hypothesize this is when the model is equipped with the basic instruction-following capability. 

\subsubsection{Distribution of Pre-Experience Data}

To better illustrate what distribution of data is required in the pre-experience process, extensive experiments are conducted to consider choosing data by ``Difficulty'', ``Diversity'' and ``Random''. In the ``Difficulty'' setting, we select $1000$ pre-experienced samples by calculating the IFD scores based on the initial raw model. In the ``Diversity'' setting, we select $1000$ data by implementing the K-means algorithm. In the ``Random'' setting, we directly select pre-experienced data randomly. After obtaining these data samples with different distributions, pre-experienced models are trained for selecting further cherry data. The performance of using $5\%$, $10\%$, and $15\%$ cherry data compared with the Alpaca model is shown in Table \ref{tbl:init}. Comparing random selection and data diversity and instruction difficulty, they all surpass the Alpaca model and are comparable to each other, indicating the effectiveness of both strategies and further proving that our IFD metric is robust across different pre-experienced models. This experiment further illustrates that what matters is this pre-experience process, rather than the sampling strategies for this process.

\vspace{-5.2mm}

\begin{table}[!t]
\centering
\scalebox{0.9}{
\begin{tabular}{lcccc}
\hline
& \textbf{5\%} & \textbf{10\%} & \textbf{15\%} & \textbf{100\%} \\\hline
Difficulty (1000)   & 1.057     & 1.072       & 1.096 & 1      \\
Diversity (1000) & 1.050     & 1.097       & 1.064 & 1      \\
Random (1000) & 1.007     & 1.047       & 1.077 & 1      \\
\hline
\end{tabular}
}
\caption{
The overall winning score changes by comparing models with different strategies of selecting pre-experienced samples with the official Alpaca model, utilizing ChatGPT. 
}
\label{tbl:init}
\vspace{-4.2mm}
\end{table}

\vspace{-2mm}
\subsection{Results on LLaMA2 Models}

In this section, experiments on newer LLaMA2-7B and LLaMA2-13B models are conducted as shown in Table \ref{tbl:llama_bench}. In these experiments, the IFD score of each sample is calculated directly based on the corresponding LLaMA2 pre-trained models by using prompts from Vicuna \cite{vicuna2023}. On both LLaMA2-7B and LLaMA2-13B models, our cherry models trained with much less data outperform the models trained with original full data. These experimental results illustrate the consistent advantages of our method and further verify the generalizability of our method. 

\begin{table*}[!htbh]
\centering
\scalebox{0.80}{
\begin{tabular}{l|ccccc|c}
\hline
& \multicolumn{5}{c|}{Huggingface Open LLM Leaderboard} & AlpacaEval \\
& Average & ARC & HellaSwag & MMLU & TruthfulQA & AlpacaEval \\
\hline
Alpaca llama2 7b & 55.25 & 54.35 & 78.65 & 47.02 & 40.98 & 27.75 \\
Ours (5\% Alpaca) & 55.78 & 57.94 & 80.37 & 44.19 & 40.62 & 36.78 \\
Ours (10\% Alpaca) & 56.31 & 58.02 & 80.42 & 46.64 & 40.18 & - \\
Ours (15\% Alpaca) & 56.37 & 57.42 & 80.68 & 46.40 & 40.95 & - \\
\hline
Alpaca llama2 13b & 58.78 & 57.59 & 81.98 & 54.05 & 41.49 & 35.00 \\
Ours (5\% Alpaca) & 61.21 & 62.37 & 84.00 & 55.65 & 42.82 & 46.82 \\
Ours (10\% Alpaca) & 61.02 & 62.97 & 83.88 & 55.29 & 41.93 & - \\
Ours (15\% Alpaca) & 61.23 & 62.37 & 83.48 & 55.56 & 43.42 & - \\
\hline
\end{tabular}
}
\caption{
The comparison of performance on Huggingface Open LLM Leaderboard and AlpacaEval Leaderboard.
}
\label{tbl:llama_bench}
\vspace{-4.2mm}
\end{table*}

\section{Cherry Data Characteristics}

\vspace{-1.2mm}%Put here to reduce too much white space after your table 

\subsection{Distribution Characteristics }
\begin{figure}[!t]
\vspace{-3.2mm}%Put here to reduce too much white space after your table
\centering 
\includegraphics[width=0.4\textwidth]{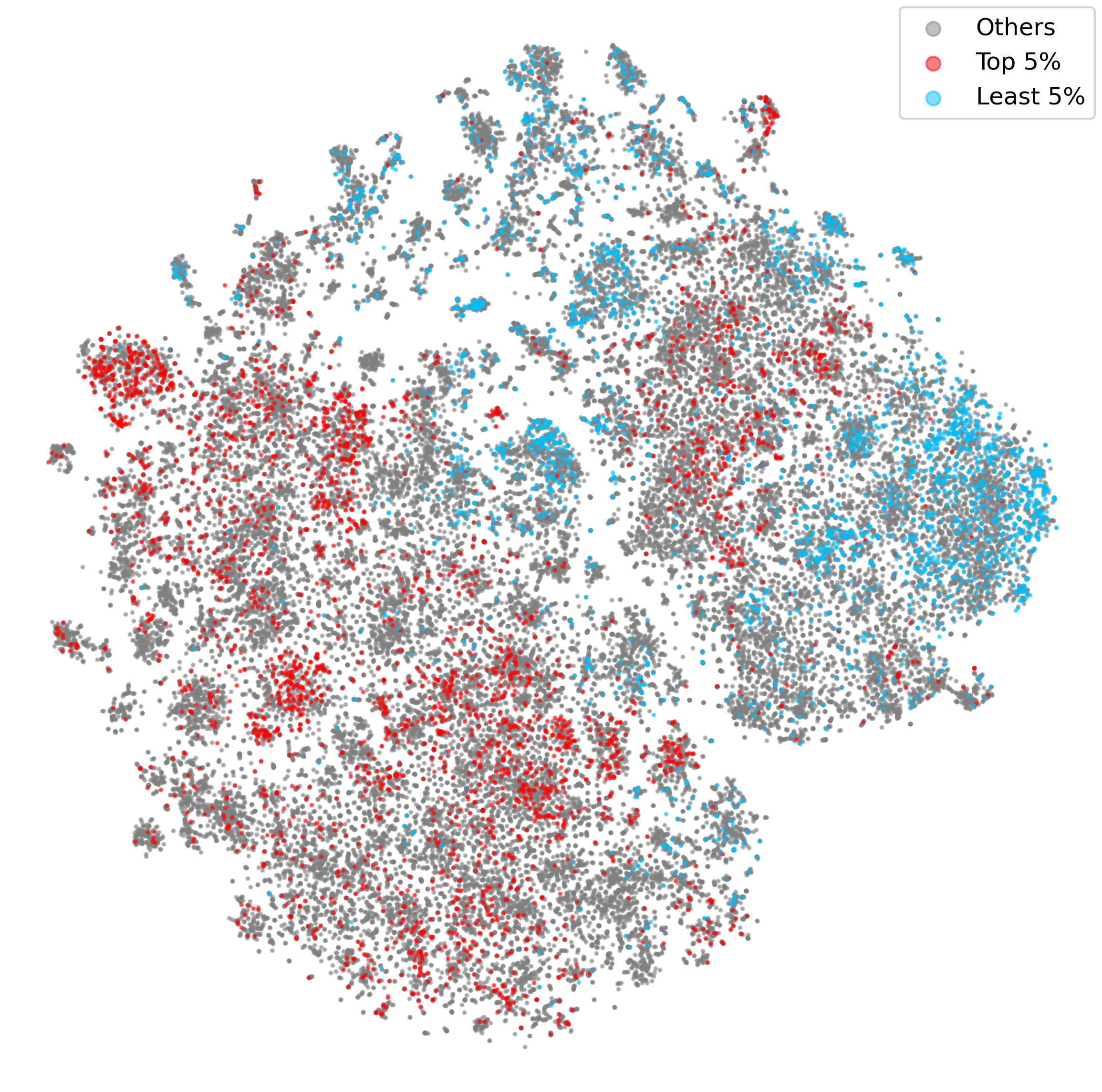} 
\vspace{-3mm}%Put here to reduce too much white space after your table 
\caption{
Visualization using t-SNE on instruction embeddings from the Alpaca dataset. Red points represent samples with the top $5\%$ IFD scores and Blue points represent samples with the least $5\%$ IFD scores. 
} 
\label{dis_top_least} 
\vspace{-6mm}
\end{figure}

In this segment, our focus is on understanding the distributional properties of the cherry data within the original dataset. Specifically, we first compute the embedding of each instruction in the Alpaca dataset and employ t-SNE for dimensionality reduction, mapping high-dimensional embeddings to 2D space. The visualized vectors, color-coded based on the top or least $5\%$ difficulty ratios, are showcased in Figure \ref{dis_top_least}. Contrary to conventional beliefs, our cherry data isn't uniformly scattered. Instead, clear boundaries exist between samples of high and low difficulty, challenging prior assumptions that selected data should span the entire instruction spectrum and maximize diversity. 

To delve deeper into the distributional intricacies of instruction embeddings, the clusters with dense high IFD scores and clusters with dense low IFD scores are manually examined. Clusters dominated by low IFD score samples are replete with rudimentary tasks like editing punctuation, words, or sentences. In contrast, high IFD score clusters are typified by deeper, more intricate tasks such as storytelling or elucidation of phenomena. We posit that these in-depth tasks are paramount for aligning large language models, compelling them to rearrange and access their intrinsic knowledge repositories. Our methodology lends partial credence to this hypothesis, leaving room for further exploration.

\subsection{Pattern Characteristics }

To better understand the pattern characteristics of the selected cherry data, we further utilize the Berkeley Neural Parser to discern the verb-noun structure within the instruction of each data sample. This analytical approach enables us to identify the main verb and its direct noun object in each instruction, providing a direct insight into what kind of instructions are prone to be assigned with higher IFD scores or lower IFD scores. This experiment is conducted based on the Alpaca data, the top $10$ occurred verb-noun pairs from the top $5\%$ IFD scores data and the least $5\%$ IFD scores data are shown in Table \ref{tbl:vn}. 

From this experiment, a clear discrepancy is revealed between the pattern characteristics of high-IFD data and low-IFD data. 
The high-IFD data mainly involves creative and complex instructions like ``write story'', ``generate list'', and ``explain concept'', which require a lot of creativity, thinking skills, and deep understanding. 
On the contrary, the low-IFD data are more about following rules and need less creativity, showing a wide range in how much thinking and creativity different tasks demand from language models. 
As a result, the reason why IFD is a valid metric for data filtering can be summarized by its ability to find the instructions that need more creativity and deep understanding.

\begin{table}[!htbh]
\centering
\scalebox{0.72}{
\begin{tabular}{ccc | ccc}
\hline
\multicolumn{3}{c|}{\textbf{Top 5\% IFD}} & \multicolumn{3}{c}{\textbf{Lease 5\% IFD}} \\
\textbf{Verb} & \textbf{Noun} & \textbf{Count} & \textbf{Verb} & \textbf{Noun} & \textbf{Count} \\
\hline
Write & Story & 119 & Rewrite & Sentence & 155 \\
Generate & Story & 98 & Edit & Sentence & 89 \\
Generate & List & 66 & Change & Sentence & 37 \\
Explain & Concept & 48 & Classify & Sentence & 36 \\
Create & Story & 44 & Convert & Sentence & 27 \\
Write & Essay & 42 & Edit & Text & 25 \\
Create & List & 28 & Translate & Sentence & 24 \\
Write & Post & 27 & Replace & Word & 16 \\
Write & Paragraph & 27 & Rearrange & Word & 15 \\
Create & Poem & 25 & Arrange & Word & 14 \\
\hline
\end{tabular}
}
\caption{
The top $10$ occurred verb-noun pairs from the top $5\%$ IFD scores data and the least $5\%$ IFD scores data. Instructions that require creativity, thinking skills, and deep understanding tend to be assigned with higher IFD scores while instructions that are more about following rules and need less creativity tend to have lower scores. 
}
\label{tbl:vn}
\vspace{-4.2mm}
\end{table}

\section{Related Work}

\subsection{Data-driven Instruction Tuning}

Previous instruction tuning collections are typically handcrafted or task-related \cite{khashabi-etal-2020-unifiedqa, ye-etal-2021-crossfit, wei2022finetuned, wang-etal-2022-super, du-etal-2022-glm, honovich-etal-2023-unnatural}, \cite{wang-etal-2023-self-instruct} utilized GPT3 \cite{NEURIPS2020_1457c0d6} to generate $52$k distinct instructions, paving the way to generating instruction data set by distilling from teacher models \cite{xu2024survey}. After the release of Meta LLaMA\cite{touvron2023llama}, the world witnessed a surge of open-sourced instruction tuning datasets and LLMs\cite{alpaca, vicuna2023, xu2023wizardlm, selfee2023, ding2023enhancing, Li2023ReflectionTuningDR, Li2024SelectiveRS}.

\subsection{Coreset Selection}

Coreset selection is pivotal in machine learning, aimed at identifying a representative subset of data points to expedite learning in various models. This approach finds its effectiveness in SVM learning \cite{tsang2005core}, K-means \cite{har2005smaller}, and logistic regression \cite{munteanu2018coresets}. 
In neural network training, recent advancements, such as those by \citet{toneva2018empirical}, explore the dynamics of data point utility during training. They find that points infrequently forgotten have minimal impact on final model accuracy. \citet{paul2021deep} demonstrate that expected loss gradient norm scores averaged over various weight initializations, effectively prune training data without significantly compromising accuracy.  \citet{mindermann2022prioritized} use Bayesian probability theory to estimate the individual impact of training points on holdout loss, refining training efficiency.

% The common thread in these works is the focus on traditional training mechanisms. Our research pivots towards optimizing core-set selection in instruction tuning. Similar to the above methods, our score estimation relies on the feature representation of the target model.

\subsection{Instruction Data Selection}

Though consensus has been made that "quality is all you need" \cite{touvron2023llama2, zhou2023lima} for instruction tuning, finding high-quality data other than through human curation is still an under-explored topic. 
Instruction Mining \cite{cao2023instruction} evaluates various indicators and applies a statistical regression model for data selection by training numerous models. 
In contrast, ALPAGASUS \cite{chen2023alpagasus} utilizes an external, fully-trained LLM (ChatGPT) to score each sample. While effective, this approach may neglect the intrinsic abilities of the base model, relying excessively on external models.
Our work aims to develop a methodology utilizing the representation feature of the target model to identify high-quality data for instruction tuning, advancing the field with a more simple and efficient approach.

\subsection{Pointwise Mutual Information}

IFD 's concept is related to Pointwise Mutual Information (PMI), a widely used metric in NLP for assessing word pair associations and contextual relevance. Both IFD and PMI aim to evaluate correlations between elements, such as questions and answers, despite employing distinct methodologies. For instance, \citet{holtzman-etal-2021-surface} leverage PMI to manage surface form competition in generative language models. They employ PMI to assess the alignment between responses and posed questions, similar to IFD's role in assessing question-answer interactions in instructional data. \citet{wiegreffe2023increasing} further deepens the understanding of PMI by proposing methods to increase the probability mass of answer choices. And numerous other PMI applications concentrate on the dialogue task \cite{mou2016sequence, zhou-etal-2019-unsupervised}. These varied applications underscore PMI's contributions to advancing natural language processing, offering valuable context for our IFD metric.

% While uniquely formulated, IFD shares a conceptual goal with Pointwise Mutual Information (PMI), a widely used metric in natural language processing for assessing word pair associations and contextual relevance. Both IFD and PMI aim to evaluate correlations between elements, such as questions and answers, despite employing distinct methodologies. For instance, \citet{holtzman-etal-2021-surface} leverage PMI to manage surface form competition in generative language models, where multiple valid responses compete for a limited probability space. They employ PMI to assess the alignment between responses and posed questions, similar to IFD's role in assessing question-answer interactions in instructional data. Additionally, they introduce domain-specific adjustments to PMI to tailor answer probabilities for specific tasks. \citet{wiegreffe2023increasing} further deepen the understanding of PMI by proposing methods to increase the probability mass on answer choices. And numerous other PMI applications concentrate on the dialogue task \cite{mou2016sequence, zhou-etal-2019-unsupervised}. These varied applications underscore PMI's contributions to advancing natural language processing, offering valuable context for our IFD metric.

\section{Conclusion}

Our study illuminates the potential of harnessing the innate capabilities of LLMs for selecting high-quality instruction tuning data that fit the model. Through our innovative self-guided approach, LLMs demonstrate the ability to discern and cherry-pick the most pertinent data samples. Central to our methodology is the Instruction-Following Difficulty score, a novel metric adept at gauging the nuanced differences between a model's autonomous outputs and expected responses. Our findings not only emphasize the importance of data quality over quantity but also underscore the potential for cost-effective LLM training. 

\section*{Limitation}

The main limitation of this method is the inconvenience of training the pre-experienced model. The concept of the Instruction-Following Difficulty score proposed by us is simple and effective, while the inconvenient pre-experienced phase makes it hard to directly put our method into usage in real-world scenarios. Though experiments on LLaMA2 models show that calculating IFD scores directly on the base LLaMA2 models also promises a good selection, we believe using the pre-experienced phase is valuable since it equips base models with the basic instruction-following ability, making the calculation of Conditioned Answer Score more reasonable. 
As a result, we believe the use of the pre-experienced phase could be a tradeoff: From the \textbf{Research Viewpoint}, using pre-experienced models is more reasonable and performs better. From the \textbf{Real-world Implementation Viewpoint}, directly using the base model is more efficient and at the same time effective as well.

\section*{Acknowledgement}
\vspace{-1mm}
This paper is supported by the Key Research and Development Program of Guangdong Province under grant No.2021B0101400003. Li and Zhou are partially supported by the Gift Fund from Adobe. The corresponding author is Jianzong Wang from Ping An Technology (Shenzhen) Co., Ltd (jzwang@188.com) and Tianyi Zhou from the University of Maryland (tianyi@umd.edu).

\bibliography{custom}

\appendix

\clearpage
\section{Implementation Details}
\label{imlement}

For experiments on the LLaMA-7B pre-trained model, our training framework aligns with protocols from Alpaca and WizardLM datasets. The Adam optimizer \cite{kingma2017adam}, with a $2\times10^{-5}$ learning rate and a batch size of $128$, steers the training across three epochs. Our pre-experienced models, however, undergo just a single epoch of training. Training on the Alpaca dataset necessitated a max input length of $512$. For WizardLM, we opted for a $1024$ input length due to hardware constraints while its original model used $2048$, which offers an inherent edge to the original model. Another challenge with WizardLM was ``AI censure'' instances. Taking a leaf from the Vicuna strategy, we filtered these samples, resulting in a streamlined WizardLM subset with $63655$ entries. Our data selection methodology was then applied to this subset. For experiments on LLaMA2-7B and LLaMA2-13b models, we utilize the instruction prompt from Vicuna \cite{vicuna2023}. Thanks to the Flash Attention mechanism \cite{dao2022flashattention}, all models on LLaMA2 use the max length of $2048$. 

\clearpage
\section{Ablation on Data Selection Mechanism}
\label{ablation_data}

\begin{table*}[!tp]
\centering
\scalebox{0.95}{
\begin{tabular}{l|ccccc|cccc}
\hline
& \multicolumn{5}{c|}{Huggingface Open LLM Leaderboard} & \multicolumn{4}{c}{Human Evaluation} \\
& Avg & ARC & HellaSwag & MMLU & TruthfulQA & Win & Tie & Lose & Winning Score \\
\hline
Ours 5\% & 52.06 & 53.92 & 79.49 & 36.51 & 38.33 & - & - & - & - \\
Random 5\% & 50.61 & 53.52 & 79.33 & 32.90 & 36.67 & 58 & 23 & 19 & 1.39 \\
Diversity 5\% & 49.48 & 53.41 & 79.29 & 29.19 & 36.04 & 61 & 21 & 18 & 1.43 \\
Low IFD 5\% & 50.77 & 53.92 & 79.09 & 34.83 & 35.25 & 87 & 8 & 5 & 1.82 \\
High CA 5\% & 47.51 & 51.45 & 75.50 & 35.41 & 26.67 & 76 & 15 & 9 & 1.67 \\
\hline
\end{tabular}
}
\caption{
The ablation performance on Huggingface Open LLM Leaderboard and Human Evaluation.  
}
\label{tbl:ablation_data}
\vspace{-4.2mm}
\end{table*}

The ablation results of different methods on the Open LLM Leaderboard and Human Evaluation results are shown in Table \ref{tbl:ablation_data}. The human evaluation configuration is aligned with the main results. The win-tie-lose counts are from the perspective of our model: It represents our model better when the Win count is greater than the Lose count.

\clearpage

\section{Performance across Sub-Categories}

To evaluate the performance variations of our model, we scrutinize the capabilities across diverse instruction tasks. To accomplish this, we compare the response of our cherry models, trained with $5\%$ Alpaca data and $10\%$ WizardLM data, to their corresponding comparing models, the official Alpaca and the reimplemented WizardLM across sub-categories in the WizardLM and Vicuna test sets, as displayed in Table \ref{tbl:sub_vic} and Table \ref{tbl:sub_wiz}. 

Our cherry model trained on Alpaca data exhibits superior or at least comparable performance to the official Alpaca model on most of the subcategories in the Vicuna and WizardLM test sets. Notably, exceptions are observed in the Math and Coding categories, corroborating the observations made by \cite{chen2023alpagasus}. We surmise that the base 7B models inherently perform sub-optimally on these two tasks, necessitating a greater volume of data samples to effectively learn the alignment.

Our cherry model trained on WizardLM data also has a better or comparable performance compared with the reimplemented WizardLM model on most of the subcategories. Specifically, Our model underperforms in Math, Code, Complex Format, and Counterfactual. The main reason our model loses in these categories is the abundance of training data for these categories in the original dataset and the supreme abilities of the original WizardLM in these tasks, which is mentioned in \cite{xu2023wizardlm}. As a consequence, when we reduce the number of data used, our model can not be trained on these data-needed categories as much as the original model, thus leading to a relatively incomparable performance. 

\begin{table*}[!tbp]
\centering
\scalebox{0.6}{
\begin{tabular}{l|c|c|c|c|c|c|c|c|c}
\hline
 & \bf Math & \bf Coding & \bf Writing & \bf Generic & \bf Knowledge & \bf Roleplay & \bf Common-sense & \bf Fermi & \bf Counterfactual \\
\hline
Alpaca & \textbf{\small 0.33} / \textbf{\small 0.33} / \textbf{\small 0.33} & \small 0.14 / \textbf{\small 0.43} / \textbf{\small 0.43} & \textbf{\small 0.60} / \small 0.20 / \small 0.20 & \textbf{\small 0.60} / \small 0.10 / \small 0.30 & \textbf{\small 0.60} / \small 0.20 / \small 0.20 & \textbf{\small 0.60} / \small 0.10 / \small 0.30 & \textbf{\small 0.60} / \small 0.20 / \small 0.30 & \textbf{\small 0.50} / \small 0.30 / \small 0.20 & \textbf{\small 0.70} / \small 0.10 / \small 0.20 \\
\hline
WizardLM & \small 0.00 / \small 0.33 / \textbf{\small 0.67} & \small 0.00 / \small 0.29 / \textbf{\small 0.71} & \textbf{\small 0.60} / \small 0.20 / \small 0.20 & \textbf{\small 0.40} / \textbf{\small 0.40} / \small 0.20 & \textbf{\small 0.70} / \small 0.30 / \small 0.00 & \textbf{\small 0.40} / \small 0.30 / \small 0.30 & \textbf{\small 0.70} / \small 0.30 / \small 0.00 & \textbf{\small 0.50} / \small 0.30 / \small 0.20 & \small 0.20 / \small 0.20 / \textbf{\small 0.60} \\
\hline
\end{tabular}
}
\caption{
The comparison between our cherry models and their corresponding comparing models on sub-categories in Vicuna test sets, using GPT4 as the judge. 
}
\label{tbl:sub_vic}
% \vspace{-4.2mm}
\end{table*}

\begin{table*}[h]
\centering
\scalebox{0.57}{
\begin{tabular}{l|c|c|c|c|c|c|c|c|c}
\hline
& \bf Math & \bf CodeGeneration & \bf Writing & \bf Computer & \bf Reasoning & \bf ComplexFormat & \bf CodeDebug & \bf CommonSense & \bf Counterfactual \\
\hline
Alpaca 
& \small 0.21 / \small 0.37 / \textbf{\small 0.42} 
& \small 0.28 / \small 0.33 / \textbf{\small 0.39} & \textbf{\small 0.56} / \small 0.17 / \small 0.28 & \textbf{\small 0.40} / \small 0.33 / \small 0.27 & \small 0.31 / \textbf{\small 0.54} / \small 0.15 & \textbf{\small 0.50} / \small 0.25 / \small 0.25 & \textbf{\small 0.50} / \textbf{\small 0.50} / \small 0.00 & \textbf{\small 0.55} / \small 0.11 / \small 0.33 & \textbf{\small 1.00} / \small 0.00 / \small 0.00 \\
\hline
WizardLM 
& \textbf{\small 0.42} / \small 0.37 / \small 0.21 
& \small 0.33 / \small 0.28 / \textbf{\small 0.39} & \textbf{\small 0.50} / \small 0.44 / \small 0.06 & \small 0.33 / \textbf{\small 0.40} / \small 0.27 & \textbf{\small 0.38} / \small 0.23 / \textbf{\small 0.38} & \small 0.25 / \small 0.25 / \textbf{\small 0.50} & \textbf{\small 0.40} / \textbf{\small 0.40} / \small 0.20 & \textbf{\small 0.56} / \small 0.44 / \small 0.00 & \small 0.00 / \small 0.38 / \textbf{\small 0.62} \\
\hline
 & \bf Multilingual & \bf Roleplay & \bf Biology & \bf Technology & \bf Ethics & \bf TruthfulQA & \bf Sport & \bf Law & \bf Medicine \\
\hline
Alpaca 
& \small 0.29 / \small 0.29 / \textbf{\small 0.42} 
& \textbf{\small 0.67} / \small 0.17 / \small 0.17 & \textbf{\small 0.50} / \small 0.00 / \textbf{\small 0.50} & \textbf{\small 0.83} / \small 0.17 / \small 0.00 & \textbf{\small 0.67} / \small 0.00 / \small 0.33 & \textbf{\small 0.60} / \small 0.00 / \small 0.40 & \textbf{\small 1.00} / \small 0.00 / \small 0.00 & \small 0.40 / \small 0.00 / \textbf{\small 0.60} & \textbf{\small 0.80} / \small 0.00 / \small 0.20 \\
\hline
WizardLM 
& \small 0.14 / \textbf{\small 0.71} / \small 0.14 
& \textbf{\small 0.33} / \textbf{\small 0.33} / \textbf{\small 0.33} 
& \small 0.17 / \textbf{\small 0.50} / \small 0.33 & \textbf{\small 0.50} / \textbf{\small 0.50} / \small 0.00 & \small 0.17 / \textbf{\small 0.83} / \small 0.00 & \textbf{\small 0.80} / \small 0.20 / \small 0.00 & \small 0.20 / \textbf{\small 0.60} / \small 0.20 & \small 0.20 / \textbf{\small 0.60} / \small 0.20 & \textbf{\small 0.80} / \small 0.00 / \small 0.20 \\
\hline
\end{tabular}
}
\caption{
The comparison between our cherry models and their corresponding comparing models on sub-categories in WizardLM test sets, using GPT4 as the judge. 
}
\label{tbl:sub_wiz}
\vspace{-4.2mm}%Put here to reduce too much white space after your table 
\end{table*}

% \clearpage
% \section{Model Training Details}

% Rooted in the Llama-7b pre-trained model, our training framework aligns with protocols from Alpaca and WizardLM datasets. The Adam optimizer \cite{kingma2017adam}, with a $2\times10^{-5}$ learning rate and a batch size of $128$, steers the training across three epochs. Our pre-experienced models, however, undergo just a single epoch of training. Training on the Alpaca dataset necessitated a max input length of $512$. For WizardLM, we opted for a $1024$ input length due to hardware constraints while its original model used $2048$, which offers an inherent edge to the original model. Another challenge with WizardLM was ``AI censure'' instances. Taking a leaf from the Vicuna strategy, we filtered these samples, resulting in a streamlined WizardLM subset with $63655$ entries. Our data selection methodology was then applied to this subset. Samples with IFD scores higher than $1$ will be filtered out before selection. For experiments on llama2-7b and llama2-13b models, we utilize the instruction prompt from Vicuna \cite{vicuna2023}. Thanks to the Flash Attention mechanism \cite{dao2022flashattention}, all models on llama2 use the max length of $2048$. 

\clearpage
\section{Results with Official WizardLM}

In this section, we provide the results of using $40\%$ of the WizardLM data to have a comparable performance with the official WizardLM model in a relatively unfair setting. The official WizardLM is uncensored and trained with the max token size of $2048$, while our model is trained with the max token size of $1024$, representing an inherent disadvantage of our model. However, even with this situation, our model can still reach a comparable performance with the official WizardLM model, inferring the effectiveness of our method. 

\begin{figure}[tbh]
\centering 
\includegraphics[width=0.5\textwidth]{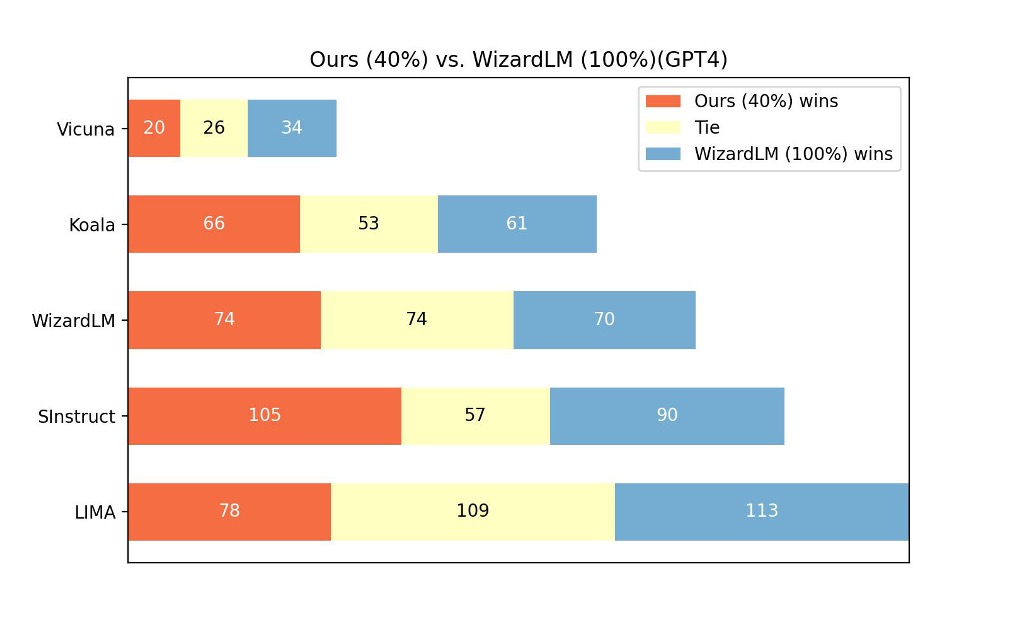} 
\caption{
Comparing our models trained on cherry data with official WizardLM trained on full data using GPT4 as the judge. 
} 
\label{appendix_wiz_40} 
\vspace{-3.2mm}%Put here to reduce too much white space after your table 
\end{figure}

\begin{table*}[!htbh]
\centering
\scalebox{1.0}{
\begin{tabular}{l|ccccc|c}
\hline
& \multicolumn{5}{c|}{Huggingface Open LLM Leaderboard} & AlpacaEval \\
& Average & ARC & HellaSwag & MMLU & TruthfulQA & AlpacaEval \\
\hline
Official WizardLM & 54.18 & 51.60 & 77.70 & 42.70 & 44.70 & 67.64 \\
Ours (40\% WizardLM) & 52.83 & 53.07 & 77.79 & 35.29 & 45.17 & 65.09 \\
\hline
\end{tabular}
}
\caption{
The comparison of performance on Huggingface Open LLM Leaderboard and AlpacaEval Leaderboard. 
}
\label{tbl:bench_40}
\vspace{-4.2mm}
\end{table*}

\clearpage
\section{Cherry Data General Characteristics}

\begin{figure}[!hbp]
\vspace{-5.2mm}%Put here to reduce too much white space after your table 
\centering 
\includegraphics[width=0.40\textwidth]{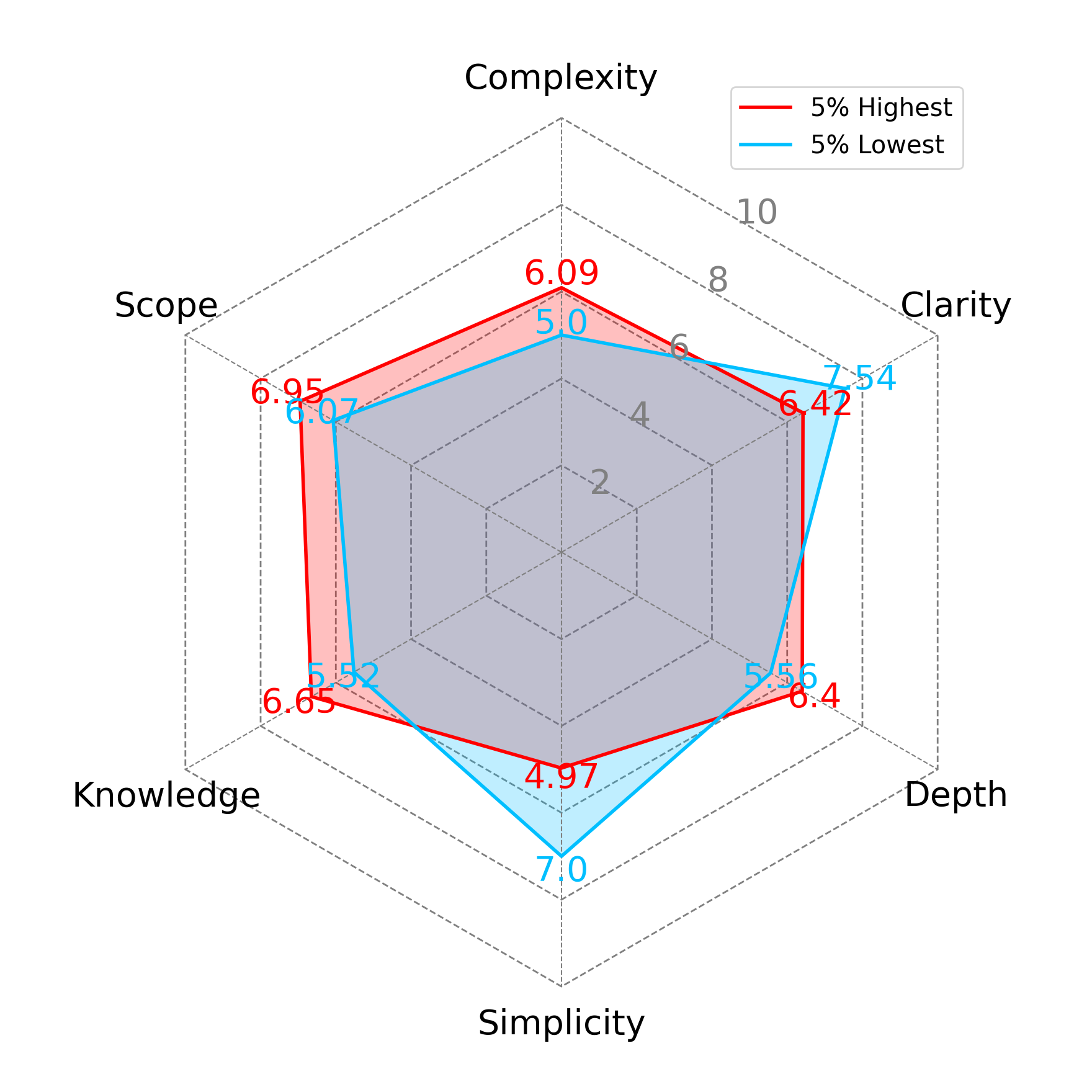} 
\vspace{-2.2mm}%Put here to reduce too much white space after your table 
\caption{
The comparison between data instances with top $5\%$ and least $5\%$ IFD scores from Alpaca data. We prompt ChatGPT to score the instruction of each data instance with respect of Scope, Complexity, Clarity, Depth, Simplicity, and Knowledge Required.  
} 
\label{hex} 
\vspace{-4.2mm}%Put here to reduce too much white space after your table 
\end{figure}

Our goal in this section is to determine if the data selected based on the IFD scores aligns with known characteristics of high-quality training data. To this end, we randomly sample $100$ instances from data with the top $5\%$ scores and the least $5\%$ scores. Utilizing ChatGPT, we evaluate each instruction on six aspects: Scope, Complexity, Clarity, Depth, Simplicity, and Knowledge Required. The results are depicted in Figure \ref{hex}. Data with a higher IFD score generally scored higher in Scope, Complexity, Depth, and Knowledge Required, but lower in Clarity and Simplicity. Simplicity, in particular, have the most pronounced discrepancy. This lends credence to our assertion that our IFD scores aptly gauge instruction complexity. Consequently, our method gravitates towards selecting more intricate samples.

As mentioned in the previous section, we try to evaluate each instruction into six aspects, Scope, Complexity, Clarity, Depth, Simplicity, and Knowledge Required. We define these aspects as follows:
\begin{itemize}
    \vspace{-2.2mm}%Put here to reduce too much white space after your table 

    \item \textbf{Scope}: The instruction encompasses the breadth and range of actions or information necessary for successful completion. 
    \vspace{-2.2mm}%Put here to reduce too much white space after your table 

    \item \textbf{Complexity}: The instruction integrates multiple steps or concepts that require careful attention and understanding.
    \vspace{-2.2mm}%Put here to reduce too much white space after your table 

    \item \textbf{Clarity}: The instruction is articulated straightforwardly, ensuring it's easily understood without ambiguity.
    \vspace{-2.2mm}%Put here to reduce too much white space after your table 

    \item \textbf{Depth}: The instruction provides thorough details and nuances, ensuring a comprehensive understanding of the task at hand.
    \vspace{-2.2mm}%Put here to reduce too much white space after your table 

    \item \textbf{Simplicity}: While thorough, the instruction avoids unnecessary jargon or convolutions, making it accessible and easy to follow.
    \vspace{-2.2mm}%Put here to reduce too much white space after your table 

    \item \textbf{Knowledge Required}: The instruction acknowledges and, if necessary, provides the foundational knowledge or context the user needs for successful execution.
\end{itemize}

From the previous Figure \ref{hex}, we can see samples selected with top IFD scores have larger scores in the aspects that reflect the difficulty of instruction, including Scope, Complexity, Depth, and Knowledge Required. These samples only underscore samples with the lowest IFD scores on the aspect of Clarity and Simplicity. This experiment detailedly illustrates the difference between samples with high or low IFD scores and verifies the effectiveness of our method in measuring the difficulty of an instruction.

\clearpage
\section{Cherry Example Analysis}

To illustrate the implications of our findings and demonstrate the characteristics of the data selected by our method, we provide several examples in Figure \ref{example}. 

The first positive example presents the situation that both the direct answer score (DA) and the conditioned answer score (CA) are relatively high. In this situation, the high DA means that it is hard for the initial pre-trained LLM to generate this poem, and the high CA means given the instruction does not make the generation of this poem much easier. So it is valuable for LLM to learn this sample. The second positive example presents the situation that both the CA score and DA score are relatively low. The low DA score means that LLM has learned this knowledge it is easy for LLM to generate this sentence. However, providing the corresponding instruction does not change the situation much, indicating the poor ability to follow this instruction.

The first negative example presents a situation where the response is too short. Due to the intrinsic nature of next token prediction that longer texts tend to have lower perplexity, the DA score is relatively high for the response that is too short and thus causes the IFD Score large, which we believe is a good feature of our method. The second negative example presents a situation where the DA score and CA score are relatively small. In this example, the response is quoted from a book that LLM must have read, thus as a known knowledge, it is easy for LLM to reproduce this sentence. However, with an instruction included, the CA score becomes even much lower, indicating LLM has gained quite a good ability in following this instruction. The third example presents the most common situation, where the instruction is simply not difficult enough.

\begin{figure*}[!t]
  \centering
  \parbox{\textwidth}{
        \rule{\textwidth}{1.5pt} % The line
        \begin{minipage}{0.47\textwidth}
            \textbf{Positive Examples}
        \end{minipage}
        \hfill
        \begin{minipage}{0.47\textwidth}
            \textbf{Negative Examples}
        \end{minipage}
        \rule{\textwidth}{0.8pt} % The line

        \begin{minipage}{0.47\textwidth}
            \textcolor{teal}{[Instruction]} Create a poem that tells the story of a struggle against an unseen force.
            
            \textcolor{teal}{[Input]} 
            
            \textcolor{teal}{[Response]} \\
            A distant enemy looms on the horizon, \\
            Selves of the light fight despite fear and fatigue. \\
            Innumerable obstacles barring their mission,\\
            Though seeming to weaken, strengths within achieve.\\

            Roots of bravery sustained their cause,\\
            Though many obstacles pushed and pulled.\\
            Heart's noise drowned out every pause,\\
            Escaping each struggle stronger and full.\\

            The reward of their labors thought clear and light,\\
            The force of the shadows still kept them at bay.\\
            Yet further pushed and tired through the night,\\
            Triumph steadily came in a growing display.\\

            Peace of their struggle found in the morning's glow,\\
            Rising higher than any unseen force's ebb and flow.

            \textcolor{teal}{[Direct Answer score (DA)]}$3.970$

            \textcolor{teal}{[Conditioned Answer score (CA)]}$3.337$
            
            \textcolor{teal}{[IFD score]}$0.928$ 

            \rule{\textwidth}{0.8pt} % The line

            \textcolor{teal}{[Instruction]} Give a brief description of the coronavirus.
            
            \textcolor{teal}{[Input]} 
            
            \textcolor{teal}{[Response]} \\
            Coronavirus is a type of virus that can cause illnesses ranging from the common cold to severe diseases such as Middle East Respiratory Syndrome (MERS) and Severe Acute Respiratory Syndrome (SARS). The virus is named for the crown-like spikes on its surface. It spreads primarily through close contact with an infected person, through small droplets from the nose or mouth after coughing or sneezing. Symptoms can include fever, tiredness, and dry cough. In severe cases, it can cause difficulty breathing, pneumonia, kidney failure, and even death.

            \textcolor{teal}{[Direct Answer score (DA)]}$0.761$

            \textcolor{teal}{[Conditioned Answer score (CA)]}$0.696$
            
            \textcolor{teal}{[IFD score]}$0.914$

        \end{minipage}
        \hfill
        \begin{minipage}{0.47\textwidth}
            \textcolor{teal}{[Instruction]} What emotion is expressed in this tweet?
            
            \textcolor{teal}{[Input]} I'm so frustrated right now!!
            
            \textcolor{teal}{[Response]} Frustration 

            \textcolor{teal}{[Direct Answer score (DA)]} $6.593$

            \textcolor{teal}{[Conditioned Answer score (CA)]} $0.601$
            
            \textcolor{teal}{[IFD score]} $0.091$ 

            \rule{\textwidth}{0.8pt} % The line

            \textcolor{teal}{[Instruction]} Identify a famous quote from the given book
            
            \textcolor{teal}{[Input]} Book: Pride and Prejudice by Jane Austen
            
            \textcolor{teal}{[Response]} "It is a truth universally acknowledged, that a single man in possession of a good fortune, must be in want of a wife." 

            \textcolor{teal}{[Direct Answer score (DA)]} $0.497$

            \textcolor{teal}{[Conditioned Answer score]} $0.026$
            
            \textcolor{teal}{[IFD score]} $0.053$

            \rule{\textwidth}{0.8pt} % The line

            \textcolor{teal}{[Instruction]} Generate a report summarizing a conversation about car maintenance.
            
            \textcolor{teal}{[Input]} 
            Joe and Lisa were discussing the importance of regular car maintenance such as oil changes, tire rotations, and brake inspects.\\
            Joe believes that regular car maintenance is essential for keeping the car operating properly and avoiding major repairs. Lisa believes that some car maintenance procedures are unnecessary and are a waste of money.
            
            \textcolor{teal}{[Response]} \\
            Joe and Lisa discussed the importance of regular car maintenance such as oil changes, tire rotations, and brake inspections. Joe believes that regular car maintenance is essential for keeping the car operating properly and avoiding major repairs while Lisa believes that some procedures are unnecessary and are a waste of money. It is agreed that regular oil changes and tire rotations are important to maintain the health of the car but that some other maintenance procedures can be skipped depending on the age and condition of the car.

            \textcolor{teal}{[Direct Answer score (DA)]} 1.667

            \textcolor{teal}{[Conditioned Answer score]} 0.599
            
            \textcolor{teal}{[IFD score]} 0.359
        \end{minipage}

        \rule{\textwidth}{0.8pt} % The line

  }
\caption{
Example instances selected by our proposed method. $2$ positive examples and $3$ negative examples are provided. 
} 
\label{example} 
\vspace{-4.2mm}%Put here to reduce too much white space after your table 
\end{figure*}

\clearpage
\section{Additional Discussion}

\subsection{Fully-trained Model as Pre-Experienced Model?}

In our method, efforts are conducted to keep the pre-experience process as simple as possible, however, there still exists a question of whether the fully-trained model can be the pre-experienced model for selecting the cherry samples. To better illustrate this question, the fully-trained Alpaca model is utilized as the pre-experienced model for selecting the cherry data, $5\%$, $10\%$, and $15\%$ of the cherry data are selected and the corresponding cherry models are trained. The performances are shown in Table \ref{tbl:fully}, in which the models with the fully-trained Alpaca hardly surpass the Alpaca with fewer data and our models. This experiment proves that the fully-trained model is not appropriate in selecting samples for the initial raw model, which is caused by the overly distribution gap between the fully-trained models and raw models. 

\begin{table}[h]
\centering
\scalebox{0.85}{
\begin{tabular}{lcccc}
\hline
& \textbf{5\%} & \textbf{10\%} & \textbf{15\%} & \textbf{100\%} \\\hline
Ours & 1.050     & 1.097       & 1.064 & 1      \\
Fully-trained Alpaca   & 0.968     & 0.999       & 1.005 & 1      \\
\hline
\end{tabular}
}
\caption{
The overall winning score changes over the data
growth comparing models with fully-trained Alpaca as the pre-experienced model with the official Alpaca model. All the comparison in this table is performed by ChatGPT.
}
\label{tbl:fully}
\vspace{-4.2mm}
\end{table}

\vspace{-1.2mm}%Put here to reduce too much white space after your table 
\subsection{How Many Cherry Samples are Required?}

While extensive experiments with our method on Alpaca and WizardLM prove the effectiveness of our method in selecting high-quality samples from the original target dataset automatically, it is still under-exploring how much data is optimal. Unlike \cite{chen2023alpagasus} in which the scores of target samples are scarce, the dense scores from our method provide better flexibility in deciding how much data you can use. However, this flexibility is also a curse that makes it hard to conclude the optimal number of data to select, which is influenced by various factors including the absolute values of the IFD scores, the distribution of hard examples, and the number of data in original datasets. However, from our empirical study, we think selecting samples with the top $10\%$ IFD scores would be a safe and reasonable choice.

\clearpage
\section{Prompt for Evaluation}

In this section, we provide the detailed prompt we used for evaluating the performance of two responses for the same instruction as shown in Figure \ref{appendix_prompt}. 

\begin{figure}[h]
  \centering
  \parbox{0.48\textwidth}{
        \rule{0.48\textwidth}{1.5pt} % The line
        Prompt for Performance Evaluation \\
        \rule{0.48\textwidth}{0.8pt} % The line
        \textbf{System Prompt} \\
        You are a helpful and precise assistant for checking the quality of the answer. \\

        \textbf{User Prompt} \\
        \text{[Question]}\\
        \textit{Question}\\
        \text{[The Start of Assistant 2's Answer]}\\
        \textit{Answer 2}\\
        \text{[The End of Assistant 2's Answer]}\\
        \text{[The Start of Assistant 2's Answer]}\\
        \textit{Answer 2}\\
        \text{[The End of Assistant 2's Answer]}\\

        We would like to request your feedback on the performance of two AI assistants in response to the user question displayed above. \\
        Please rate the helpfulness, relevance, accuracy, level of details of their responses. Each assistant receives an overall score on a scale of 1 to 10, where a higher score indicates better overall performance. \\
        Please first output a single line containing only two values indicating the scores for Assistant 1 and 2, respectively. The two scores are separated by a space. In the subsequent line, please provide a comprehensive explanation of your evaluation, avoiding any potential bias and ensuring that the order in which the responses were presented does not affect your judgment.

        \rule{0.48\textwidth}{0.8pt} % The line

  }
\caption{
The prompt we used to request ChatGPT or GPT4 to evaluate the responses. 
} 
\label{appendix_prompt} 
\vspace{-4.2mm}%Put here to reduce too much white space after your table 
\end{figure}

\clearpage
\section{Detailed Main Comparison}

\subsection{Comparison with the Official Alpaca}

\begin{figure*}[tbh]
\centering 
\includegraphics[width=0.98\textwidth]{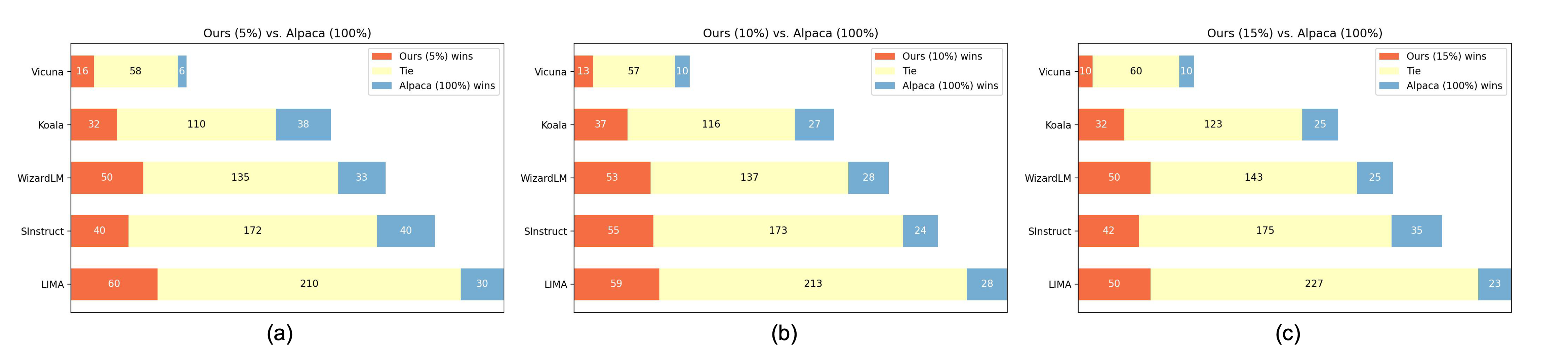} 
\caption{
Comparing our cherry models with the official Alpaca model from $5\%$ to $15\%$ of the data using ChatGPT as the judge. Each horizontal bar represents a comparison in a specific test set. 
} 
\label{appendix_compare_alpaca} 
\vspace{-4.2mm}%Put here to reduce too much white space after your table 
\end{figure*}

As shown in Figure \ref{appendix_compare_alpaca}, we present the detailed comparison between our cherry models with the official Alpaca (7B) model across different test set with different percentage of cherry data, from $5\%$ to $15\%$, using ChatGPT as the judge. Starting from $5\%$ of the full data, our cherry models outperform the official Alpaca model in all these data scales. 

\subsection{Comparison with the Reimplemented WizardLM }

\begin{figure*}[tbh]
\centering 
\includegraphics[width=0.98\textwidth]{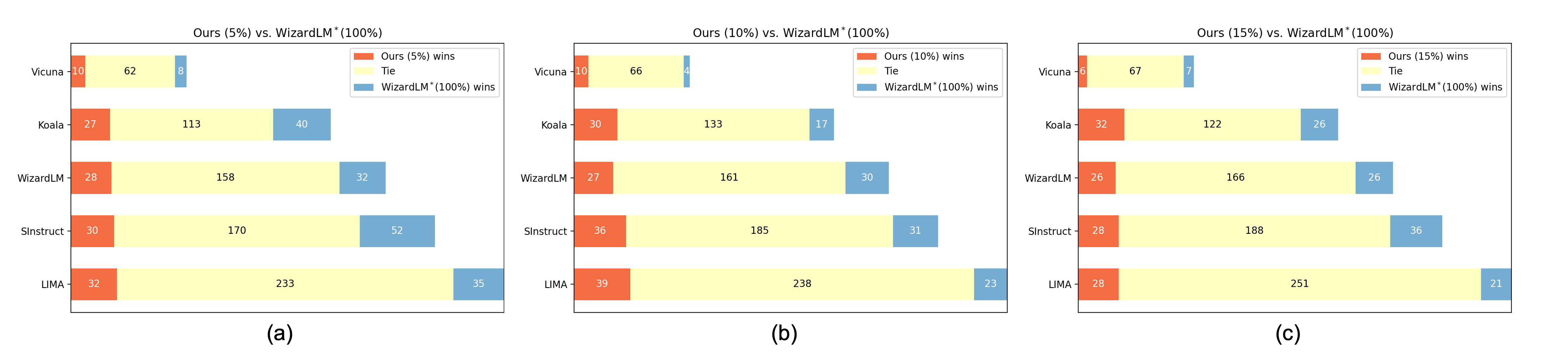} 
\caption{
Comparing our cherry models with the reimplemented WizardLM model from $5\%$ to $15\%$ of the data using ChatGPT as the judge. Each horizontal bar represents a comparison in a specific test set. 
} 
\label{appendix_compare_wiz_1} 
\vspace{-4.2mm}%Put here to reduce too much white space after your table 
\end{figure*}

As shown in Figure \ref{appendix_compare_wiz_1}, we present the detailed comparison between our cherry models with the reimplemented WizardLM (7B) model across different test set with different percentage of cherry data, from $5\%$ to $15\%$, using ChatGPT as the judge. Our cherry models begin outperforming the reimplemented WizardLM from the scale of $10\%$ of the data. 

\subsection{Comparison with the Official WizardLM }

\begin{figure*}[tbh]
\centering 
\includegraphics[width=0.98\textwidth]{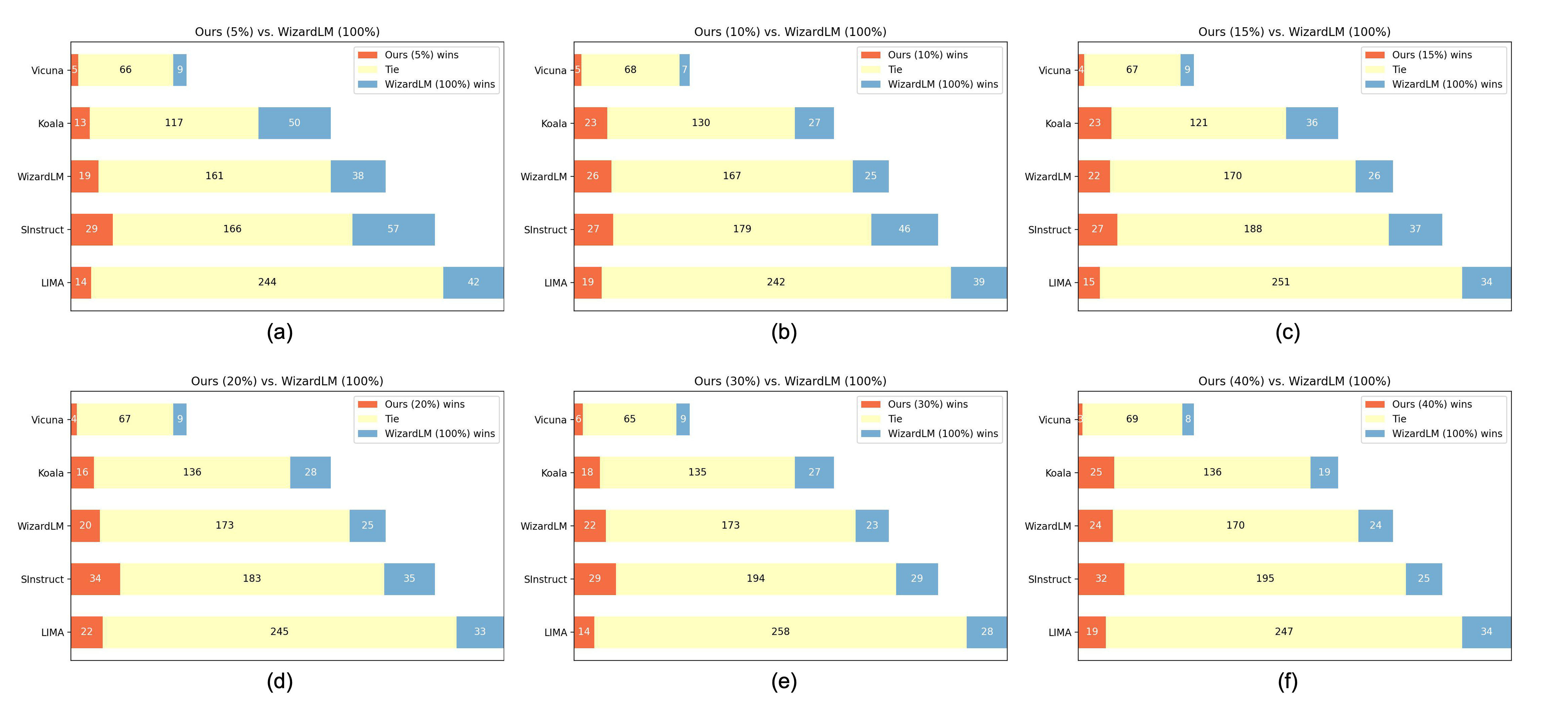} 
\caption{
Comparing our cherry models with the official WizardLM model from $5\%$ to $40\%$ of the data using ChatGPT as the judge. Each horizontal bar represents a comparison in a specific test set. 
} 
\label{appendix_compare_wiz_2} 
\vspace{-4.2mm}%Put here to reduce too much white space after your table 
\end{figure*}

As shown in Figure \ref{appendix_compare_wiz_2}, we show the detailed comparison between our cherry models with the reimplemented WizardLM (7B) model across different test set with different percentage of cherry data, from $5\%$ to $40\%$, using ChatGPT as the judge. When compared with the official WizardLM data, our cherry model achieves a comparable performance when using $40\%$ of the WizardLM data, which is positive considering the inherent disadvantage of our training configuration. 

\clearpage
\section{Detailed Ablation Comparison}

\subsection{Data Randomly Selected}

\begin{figure*}[tbh]
\centering 
\includegraphics[width=0.98\textwidth]{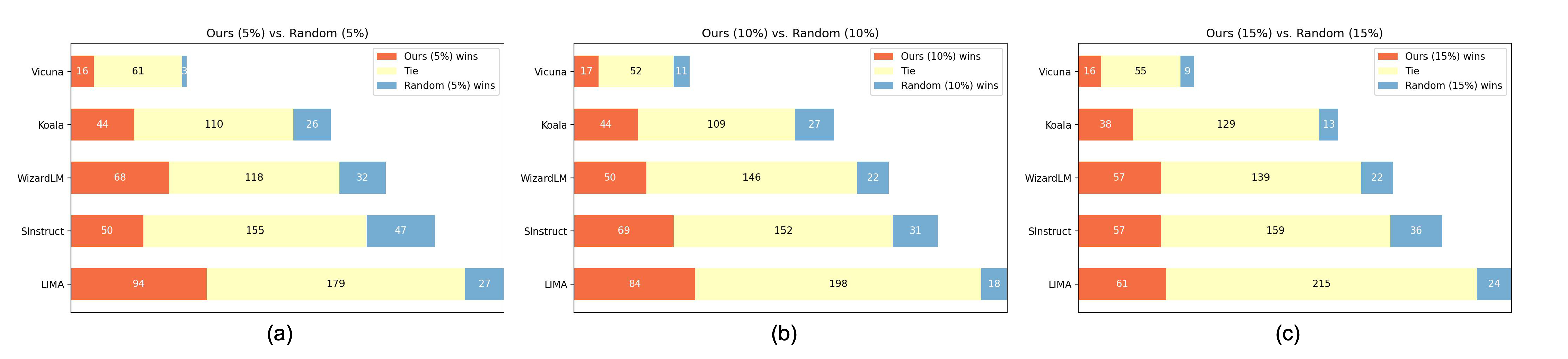} 
\caption{
Comparing our cherry models with models utilizing randomly selected data from $5\%$ to $15\%$, using ChatGPT as the judge. Each horizontal bar represents a comparison in a specific test set. 
} 
\label{appendix_compare_ablation_random} 
\vspace{-4.2mm}%Put here to reduce too much white space after your table 
\end{figure*}

As shown in Figure \ref{appendix_compare_ablation_random}(a)(b)(c), we show the detailed comparison between the models trained with randomly selected data with our cherry models across different test set with different percentage of data, from $5\%$ to $15\%$, using ChatGPT as the judge. From $5\%$ to $15\%$ of the data, our cherry models consistently outperform the random models. 

\subsection{Data with Low IFD Score}

\begin{figure*}[tbh]
\centering 
\includegraphics[width=0.98\textwidth]{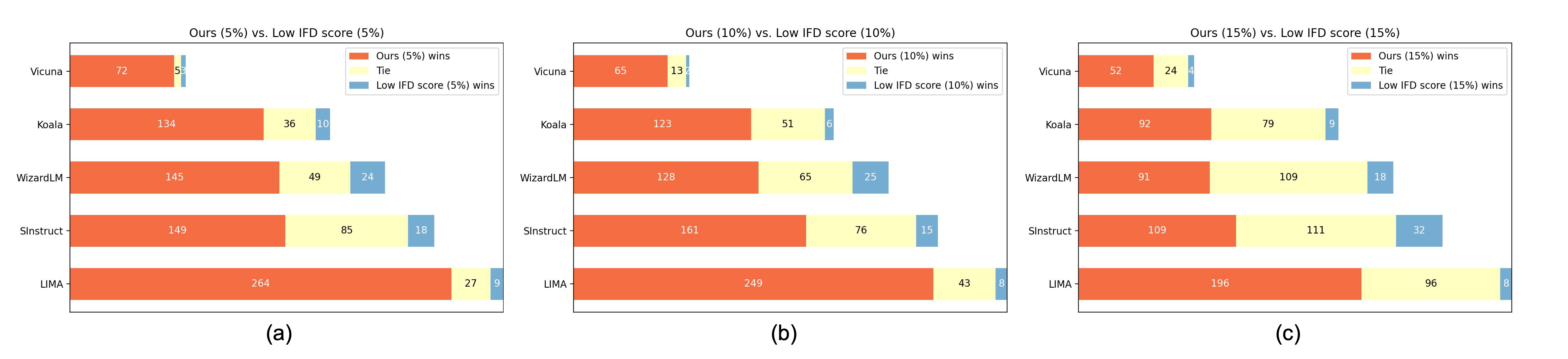} 
\caption{
Comparing our cherry models with models trained with data selected with low IFD score from $5\%$ to $15\%$, using ChatGPT as the judge. Each horizontal bar represents a comparison in a specific test set. 
} 
\label{appendix_compare_ablation_low_idf} 
\vspace{-4.2mm}%Put here to reduce too much white space after your table 
\end{figure*}

As shown in Figure \ref{appendix_compare_ablation_low_idf}, we show the detailed comparison between the models trained with data selected with low IFD scores with our cherry models across different test set with different percentage of data, from $5\%$ to $15\%$, using ChatGPT as the judge. From $5\%$ to $15\%$ of the data, our cherry models consistently have better performances.

\subsection{Data with High CA Scores}

\begin{figure*}[tbh]
\centering 
\includegraphics[width=0.98\textwidth]{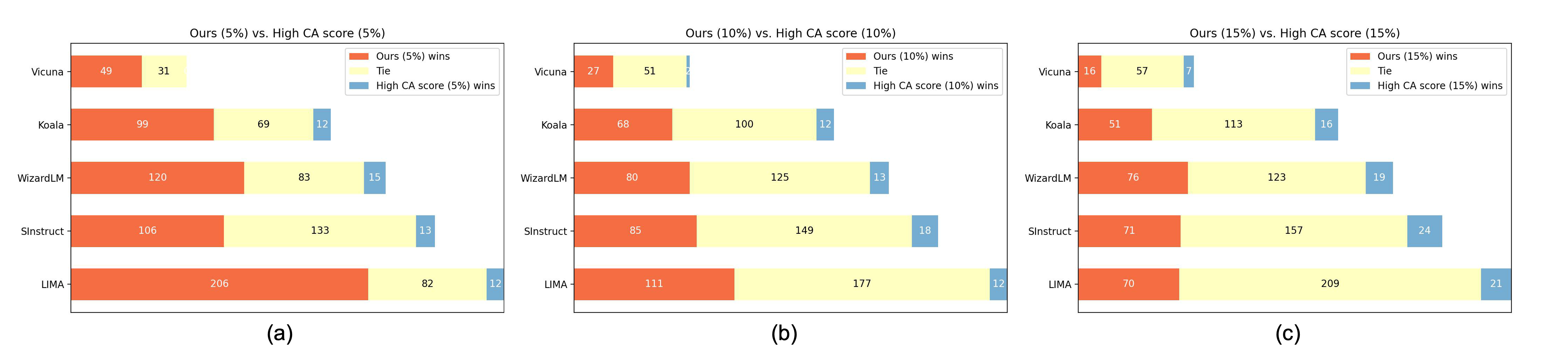} 
\caption{
Comparing our cherry models with models trained with data selected with high conditioned answer scores from $5\%$ to $15\%$, using ChatGPT as the judge. Each horizontal bar represents a comparison in a specific test set. 
} 
\label{appendix_compare_ablation_high_ca} 
\vspace{-4.2mm}%Put here to reduce too much white space after your table 
\end{figure*}

As shown in Figure \ref{appendix_compare_ablation_high_ca}, we show the detailed comparison between the models trained with data selected with high conditioned answer scores with our cherry models across different test set with different percentage of data, from $5\%$ to $15\%$, using ChatGPT as the judge. From $5\%$ to $15\%$ of the data, our cherry models consistently have better performances.

\subsection{Number of Pre-Experienced Data}

\begin{figure*}[tbh]
\centering 
\includegraphics[width=0.98\textwidth]{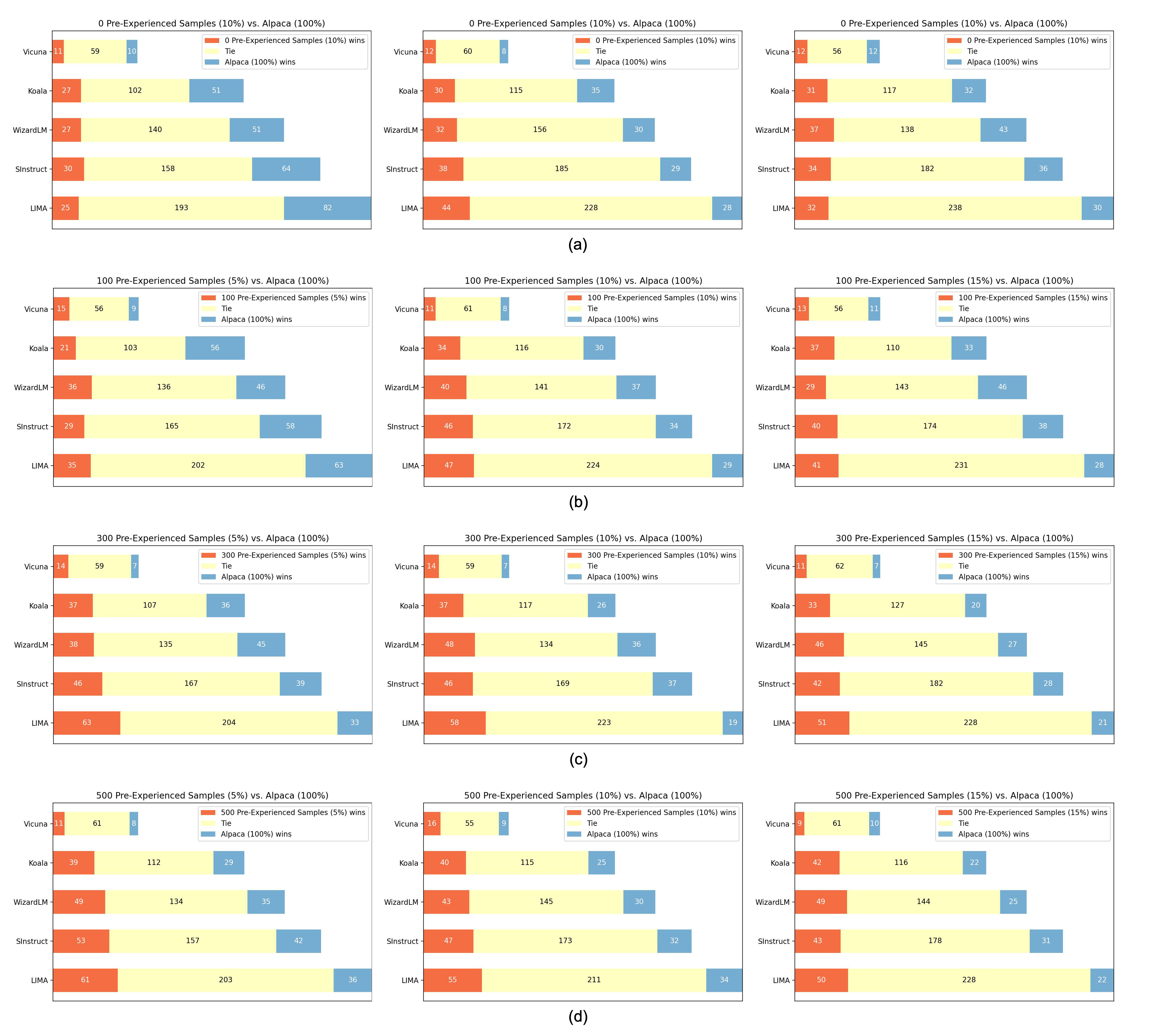} 
\caption{
Comparing our cherry models with different numbers of pre-experienced samples with the Alpca model from $5\%$ to $15\%$, using ChatGPT as the judge. (a)(b)(c) and (d) represent the situation where $0$, $100$, $300$ and $500$ pre-experienced samples are utilized respectively. 
} 
\label{appendix_pre} 
\vspace{-4.2mm}%Put here to reduce too much white space after your table 
\end{figure*}

Figure \ref{appendix_pre} shows the comparisons when different numbers of pre-experienced samples are utilized to train the pre-experienced model. 

\subsection{Distribution of Pre-Experience Data}

\begin{figure*}[tbh]
\centering 
\includegraphics[width=0.98\textwidth]{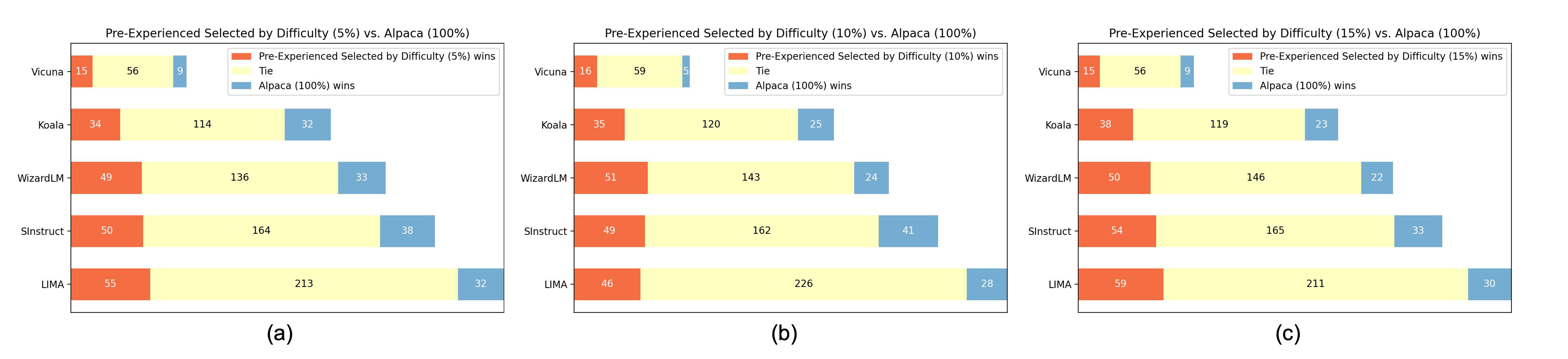} 
\caption{
Comparing models whose pre-experienced data are selected by considering the difficulty of instructions with the Alpca model from $5\%$ to $15\%$, using ChatGPT as the judge. 
} 
\label{appendix_init} 
\vspace{-4.2mm}%Put here to reduce too much white space after your table 
\end{figure*}

Figure \ref{appendix_init} shows the comparisons when IFD scores are used as the strategy to select pre-experienced data to train the pre-experienced model. 

\subsection{Fully-trained Model as Pre-Experienced Models}

\begin{figure*}[tbh]
\centering 
\includegraphics[width=0.98\textwidth]{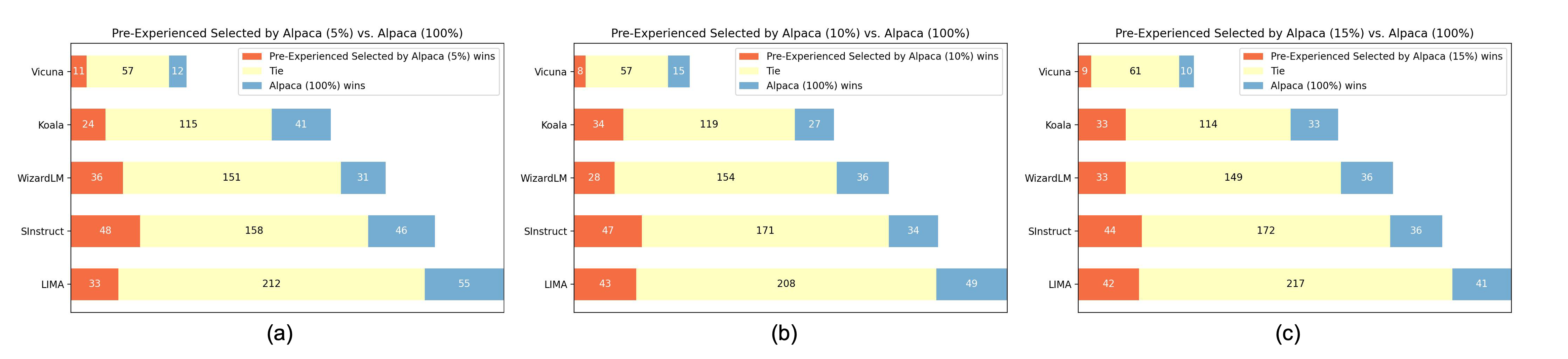} 
\caption{
Comparing models with fully-trained Alpaca as the pre-experienced model with the Alpca model from $5\%$ to $15\%$, using ChatGPT as the judge. 
} 
\label{appendix_fully} 
\vspace{-4.2mm}%Put here to reduce too much white space after your table 
\end{figure*}

Figure \ref{appendix_fully} shows the detailed comparisons when the fully-trained official Alpaca is utilized as the pre-experienced model for selecting cherry data. 

\clearpage
\section{More Examples}

In this section, some positive examples with top $5\%$ IFD scores in the Alpaca dataset are presented in Figure \ref{append_example_positive_1} and \ref{append_example_positive_2}. Negative examples with the least $5\%$ IFD scores are presented in Figure \ref{append_example_negative_1}. 

\begin{figure*}[!t]
  \centering
  \parbox{\textwidth}{
        \rule{\textwidth}{1.5pt} % The line
        \begin{minipage}{0.47\textwidth}
            \textbf{Positive Example}
        \end{minipage}
        \hfill
        \begin{minipage}{0.47\textwidth}
            \textbf{}
        \end{minipage}
        \rule{\textwidth}{0.8pt} % The line

        \begin{minipage}{0.47\textwidth}
            \textcolor{teal}{[Instruction]} Design a wellness plan for the given audience
            
            \textcolor{teal}{[Input]} Expectant Mothers
            
            \textcolor{teal}{[Response]} 
            
            This wellness plan for expectant mothers should include strategies for physical, emotional and mental health. Physical health should include exercise, a healthy diet and adequate rest. Expectant mothers should strive for at least 30 minutes of moderate exercise such as walking or swimming every day. A nutritious diet should include a variety of fruits, vegetables and proteins. Adequate sleep is essential for the body and mind to recharge, so make sure to get seven to eight hours per night. \\
            Stress management is important for emotional and mental health. This can include mindfulness activities, journaling or talking to a friend. Participating in activities like yoga, meditation or guided visualization can also be beneficial. Make sure to take time every day to treat yourself, whether it be by getting a massage or simply taking a long bubble bath.\\
            Make sure to take prenatal vitamins as prescribed by your doctor and attend all of your obstetrician checkups. It's also important to stay hydrated by drinking plenty of water. Schedule regular check-ins with your doctor to ensure the health of both you and your baby.
            
            \rule{\textwidth}{0.8pt} % The line

            \textcolor{teal}{[Instruction]} Identify the types of bias in machine learning
            
            \textcolor{teal}{[Input]} 
            
            \textcolor{teal}{[Response]} 

            The types of bias in machine learning include sampling bias, data bias, selection bias, measurement bias, model bias and inference bias. Sampling bias occurs when the dataset is not representative of the population;data bias happens when data is skewed or incomplete; selection bias arises when data is chosen in a non-random way; measurement bias is caused by incorrect labels or noisy data; model bias arises when the model is overly reliant on specific features or data; and inference bias is caused by the model being too simplistic.

            % \rule{\textwidth}{0.8pt} % The line

            % \textcolor{teal}{[Instruction]} Find one historical event in the 15th century (1400s) and describe it briefly.
            
            % \textcolor{teal}{[Input]} 
            
            % \textcolor{teal}{[Response]} 
            
            % In the 15th century, there was an event that had a major impact on the world - the fall of Constantinople. On May 29th, 1453, the Ottoman army defeated the Byzantine Empire and captured Constantinople, which led to the end of the Roman Empire and a major shift in the geopolitical landscape. Constantinople, once a major center of trade and cultural exchange, fell to the Ottoman Turks and turned into an Islamic city. This marked the beginning of a new era in the globalization of trade and the spread of ideas. The fall of Constantinople was one of the most influential events of the 15th century and is still remembered to this day. 

        \end{minipage}
        \hfill
        \begin{minipage}{0.47\textwidth}
            \textcolor{teal}{[Instruction]} Generate a story about a journey in space.
            
            \textcolor{teal}{[Input]} 
            
            \textcolor{teal}{[Response]} 

            It was a beautiful day on Earth, but Andrew was not content to stay on land. He wanted to explore and experience the vastness of space. So, on that fateful day, he set off on an intergalactic voyage to the farthest reaches of the stars. \\
            ...\\
            Exploring this new universe felt like a dream, and Andrew found himself discovering more wonders than he had ever imagined. One of his greatest adventures was traveling to a star that had never been seen by anyone before him. \\
            Years passed, and Andrew eventually returned to Earth, full of stories and memories of his travels. He had learned a great deal about the universe and made many friends along the way. His journey had taught him that space is full of boundless possibilities and wonders, and he will never forget his time in the stars.

            \rule{\textwidth}{0.8pt} % The line

            \textcolor{teal}{[Instruction]} Provide an analysis of the impact of carbohydrates on human health.
            
            \textcolor{teal}{[Input]} 
            
            \textcolor{teal}{[Response]} \\
            Carbohydrates are an important part of a balanced human diet and have been linked to a variety of health benefits when consumed in appropriate amounts. Carbohydrates are the main source of energy for the body and are important for the brain, organs, and muscles to function properly. ... Choosing complex carbohydrates, such as fruits and vegetables, whole grains, and legumes, instead of simple carbohydrates that are found in processed and refined foods, can help to provide essential vitamins, minerals and other nutrients. Ultimately, carbohydrates are a necessary part of a healthy diet and should be consumed in moderation and in a balanced way.

        \end{minipage}

        \rule{\textwidth}{0.8pt} % The line

  }
\caption{
Positive examples sampled from the alpaca dataset with top $5\%$ IFD scores. 
} 
\label{append_example_positive_1} 
\vspace{-5.2mm}%Put here to reduce too much white space after your table 
\end{figure*}

\begin{figure*}[!t]
  \centering
  \parbox{\textwidth}{
        \rule{\textwidth}{1.5pt} % The line
        \begin{minipage}{0.47\textwidth}
            \textbf{Positive Example}
        \end{minipage}
        \hfill
        \begin{minipage}{0.47\textwidth}
            \textbf{}
        \end{minipage}
        \rule{\textwidth}{0.8pt} % The line

        \begin{minipage}{0.47\textwidth}
            \textcolor{teal}{[Instruction]} Outline the steps a startup would need to take to increase brand awareness
            
            \textcolor{teal}{[Input]} 
            
            \textcolor{teal}{[Response]} 
            
            Steps to Increase Brand Awareness for a Startup: \\
            1. Identify target customers: Understand the target demographic and research their needs and interests. \\
            2. Develop a content strategy: Use content to engage, inform or entertain customers in order to establish a connection with them.\\
            3. Share the story: Create a website and create social media platforms to share the story, mission or purpose of the startup with customers. \\
            4. Make it interactive: Encourage audience participation via competitions, surveys, or polls.\\
            5. Track and analyze data: Measure the response of customers and use the insights to inform any changes or adjustments."
            
            \rule{\textwidth}{0.8pt} % The line

            \textcolor{teal}{[Instruction]} Create a poem that is related to the season of Fall.
            
            \textcolor{teal}{[Input]} 
            
            \textcolor{teal}{[Response]} 

            The leaves are falling, leaves of autumn gold,\\
            A gentle breeze, a chill that blows,\\
            The sun that slowly sets, before night comes to fold,\\
            A season of beauty, with dreams to behold.\\
            The days grow shorter, the sky turns gray,\\
            The land it sighs as summer slips away,\\
            Orange, yellow, and crimson begin to display,\\
            Bringing in a vibrant beauty, during this time of day.\\
            The winds blow cold, as Frost begins to bite,\\
            The air whispers, chill and full of fright,\\
            It is the season of change, and so we all must write,\\
            The story of our lives, in the midst of this night."

            % \rule{\textwidth}{0.8pt} % The line

            % \textcolor{teal}{[Instruction]} Brainstorm some activities that could make an in-person work meeting more engaging.
            
            % \textcolor{teal}{[Input]} 
            
            % \textcolor{teal}{[Response]} 
            
            % -Start the meeting with an icebreaker to get everyone talking \\
            % -Divide the group into smaller teams and have them work on a problem together \\
            % -Set a timer for each person to give their feedback \\
            % -Introduce short games or challenges that get everyone involved \\
            % -Take a break in the middle of the meeting and have everyone do an energizing activity \\
            % -Encourage participation and ask everyone\u2019s opinion \\
            % -End the meeting with a thank you and follow up tasks."

        \end{minipage}
        \hfill
        \begin{minipage}{0.47\textwidth}
            \textcolor{teal}{[Instruction]} What are the major components of the Java programming language?
            
            \textcolor{teal}{[Input]} 
            
            \textcolor{teal}{[Response]} 

            The major components of the Java programming language are the Java Virtual Machine (JVM), the Java Runtime Environment (JRE), the Java Platform, the Java Development Kit (JDK), and the Java Libraries. These components provide core components such as threading, garbage collection, memory management and reflection, as well as APIs such as the I/O package, networking, graphics, and more. The JDK is the most fundamental component and is required to compile Java programs. It contains a compiler, interpreter, and other tools needed to create Java applications. The JRE, on the other hand, contains the necessary components to execute the Java programs. The JVM takes the compiled Java programs and executes them. Lastly, the Java Libraries are a vast collection of software packages written in Java that provide the necessary components for programmers to develop applications.

            \rule{\textwidth}{0.8pt} % The line

            \textcolor{teal}{[Instruction]} Given a historical event, narrate an account of it in one paragraph.
            
            \textcolor{teal}{[Input]} The sinking of the Titanic
            
            \textcolor{teal}{[Response]} \\
            The sinking of the Titanic was one of the most devastating maritime disasters in history. On April 15th, 1912, the illustrious luxury liner set sail on its maiden voyage from Southampton, England, bound for New York City. Despite being touted as the \"unsinkable\" ship, the Titanic struck an iceberg and sunk in the early morning hours of the 15th. It took over two hours for the ship to completely disappear beneath the surface of the Atlantic Ocean. Over 1,500 souls were lost, and the disaster shocked the world. To this day, the tragedy of the Titanic serves as a somber reminder of the power of the sea and the fragility of human life.

            % \rule{\textwidth}{0.8pt} % The line

            % \textcolor{teal}{[Instruction]} Explain how the given topics are connected.
            
            % \textcolor{teal}{[Input]} Chemistry and Biology
            
            % \textcolor{teal}{[Response]} \\
            % Chemistry and biology are closely intertwined fields of study. Chemistry helps to explore the chemical processes necessary for life to exist, such as photosynthesis. It also  serves as a foundation for studying and understanding the molecules involved in metabolism and genetic processes. On the other hand, biology uses the knowledge of chemistry to understand the structure and function of living organisms, from single cells to complex organisms such as humans. Therefore, chemistry and biology are closely intertwined, with one aiding in the understanding of the other.

        \end{minipage}

        \rule{\textwidth}{0.8pt} % The line

  }
\caption{
Positive examples sampled from the alpaca dataset with top $5\%$ IFD scores. 
} 
\label{append_example_positive_2} 
\vspace{-5.2mm}%Put here to reduce too much white space after your table 
\end{figure*}

\begin{figure*}[!t]
  \centering
  \parbox{\textwidth}{
        \rule{\textwidth}{1.5pt} % The line
        \begin{minipage}{0.47\textwidth}
            \textbf{Negative Example}
        \end{minipage}
        \hfill
        \begin{minipage}{0.47\textwidth}
            \textbf{}
        \end{minipage}
        \rule{\textwidth}{0.8pt} % The line

        \begin{minipage}{0.47\textwidth}
            \textcolor{teal}{[Instruction]} What are the three primary colors?
            
            \textcolor{teal}{[Input]} 
            
            \textcolor{teal}{[Response]} 
            
            The three primary colors are red, blue, and yellow.
            
            \rule{\textwidth}{0.8pt} % The line

            \textcolor{teal}{[Instruction]} Determine if the following sentence is in the simple or the perfect tense: I have celebrated my birthday.
            
            \textcolor{teal}{[Input]} 
            
            \textcolor{teal}{[Response]} 

            The sentence is in the perfect tense.

            \rule{\textwidth}{0.8pt} % The line

            \textcolor{teal}{[Instruction]} Evaluate this sentence for spelling and grammar mistakes
            
            \textcolor{teal}{[Input]} He finnished his meal and left the resturant
            
            \textcolor{teal}{[Response]} \\
            He finished his meal and left the restaurant.

            \rule{\textwidth}{0.8pt} % The line

            \textcolor{teal}{[Instruction]} Rewrite the following sentence using active voice.
            
            \textcolor{teal}{[Input]} The news report was read by the captain.
            
            \textcolor{teal}{[Response]} \\
            The captain read the news report.

            \rule{\textwidth}{0.8pt} % The line

            \textcolor{teal}{[Instruction]} Based on the description, determine if the following text is positive or negative sentiment.
            
            \textcolor{teal}{[Input]} It was a disastrous event.
            
            \textcolor{teal}{[Response]} \\
            Negative sentiment

            \rule{\textwidth}{0.8pt} % The line

            \textcolor{teal}{[Instruction]} What is the force on a 1 kg mass due to the gravitational force?
            
            \textcolor{teal}{[Input]} 
            
            \textcolor{teal}{[Response]} \\
            The force on a 1 kg mass due to the gravitational force is 9.8 N.

            \rule{\textwidth}{0.8pt} % The line

            \textcolor{teal}{[Instruction]} Transform the following sentence into the passive voice
            
            \textcolor{teal}{[Input]} I bought a book
            
            \textcolor{teal}{[Response]} \\
            A book was bought by me.

            % \rule{\textwidth}{0.8pt} % The line

            % \textcolor{teal}{[Instruction]} What is the most commonly shared type of content on social media?
            
            % \textcolor{teal}{[Input]} 
            
            % \textcolor{teal}{[Response]} \\
            % The most commonly shared type of content on social media is images.

            % \rule{\textwidth}{0.8pt} % The line

            % \textcolor{teal}{[Instruction]} Edit the following sentence to correct the grammar and syntax errors
            
            % \textcolor{teal}{[Input]} You was late to the party
            
            % \textcolor{teal}{[Response]} \\
            % You were late to the party.

        \end{minipage}
        \hfill
        \begin{minipage}{0.47\textwidth}
            \textcolor{teal}{[Instruction]} Edit the given text so it reads correctly in terms of grammar, sentence structure and spelling.
            
            \textcolor{teal}{[Input]} Me and my friend went to the cinema last niught and we saw a realy great movei.
            
            \textcolor{teal}{[Response]} \\
              My friend and I went to the cinema last night and we saw a really great movie.

            \rule{\textwidth}{0.8pt} % The line

            \textcolor{teal}{[Instruction]} Rewrite the following sentence in a more concise way: "The dog that I saw in the park, it was black in color."
            
            \textcolor{teal}{[Input]} The dog that I saw in the park, it was black in color.
            
            \textcolor{teal}{[Response]} \\
            I saw a black dog in the park.

            \rule{\textwidth}{0.8pt} % The line

            \textcolor{teal}{[Instruction]} Categorize the following text into one of the two sentiment classes: positive or negative.
            
            \textcolor{teal}{[Input]} I didn't like the product very much because it was too slow and not user friendly.
            
            \textcolor{teal}{[Response]} \\
            Negative.

            \rule{\textwidth}{0.8pt} % The line

            \textcolor{teal}{[Instruction]} What type of capital did the company have?
            
            \textcolor{teal}{[Input]} The company received an equity funding of \$5 million.
            
            \textcolor{teal}{[Response]} \\
            The company had equity capital.

            \rule{\textwidth}{0.8pt} % The line

            \textcolor{teal}{[Instruction]} Name the three main mountain ranges in the United States.
            
            \textcolor{teal}{[Input]} 
            
            \textcolor{teal}{[Response]} \\
            The three main mountain ranges in the United States are the Rocky Mountains, the Appalachian Mountains, and the Sierra Nevada Mountains.

            \rule{\textwidth}{0.8pt} % The line

            \textcolor{teal}{[Instruction]} Edit the following sentence so it uses the word \"licentious\": Bill and Ted's behavior was inappropriate.
            
            \textcolor{teal}{[Input]} 
            
            \textcolor{teal}{[Response]} \\
            Bill and Ted's licentious behavior was inappropriate.

            % \rule{\textwidth}{0.8pt} % The line

            % \textcolor{teal}{[Instruction]} Identify the main character in the following story.
            
            % \textcolor{teal}{[Input]} Once upon a time, there lived a poor peasant boy who worked very hard on his farm.
            
            % \textcolor{teal}{[Response]} \\
            % The main character in the story is the poor peasant boy.
            
            % \rule{\textwidth}{0.8pt} % The line

            % \textcolor{teal}{[Instruction]} Delete the adjective 'bustling' from the following sentence and rewrite it.
            
            % \textcolor{teal}{[Input]} The bustling city of Tokyo was awe-inspiring.
            
            % \textcolor{teal}{[Response]} \\
            % The city of Tokyo was awe-inspiring.
            
        \end{minipage}

        \rule{\textwidth}{0.8pt} % The line

  }
\caption{
Negative examples sampled from the alpaca dataset with the least $5\%$ IFD scores. 
} 
\label{append_example_negative_1} 
\vspace{-5.2mm}%Put here to reduce too much white space after your table 
\end{figure*}

\end{document}